%% file: main.tex
\documentclass[10pt,twocolumn,letterpaper]{article}

\usepackage[pagenumbers]{iccv} 


\definecolor{iccvblue}{rgb}{0.21,0.49,0.74}
\usepackage[pagebackref,breaklinks,colorlinks,allcolors=iccvblue]{hyperref}
\usepackage{paralist}

\usepackage[utf8]{inputenc} 
\usepackage[T1]{fontenc}    
\usepackage{amsfonts}       
\usepackage{nicefrac}       
\usepackage{microtype}      
\usepackage{xcolor}         
\usepackage[linesnumbered,ruled,vlined]{algorithm2e}
\usepackage{array}
\usepackage{amssymb}
\usepackage{latexsym}
\usepackage{amsmath,amssymb}
\usepackage{graphicx}
\usepackage{multirow}
\usepackage{adjustbox}
\usepackage{makecell}
\usepackage[nottoc]{tocbibind}
\usepackage{graphicx}
\usepackage{bm}
\usepackage{kotex}
\usepackage{booktabs, multirow} 
\usepackage{nicematrix,caption}
\usepackage{soul}
\usepackage{changepage,threeparttable} 
\usepackage{subcaption}

\def\paperID{7698} 
\def\confName{ICCV}
\def\confYear{2025}

\title{PersonaCraft: Personalized and Controllable Full-Body Multi-Human Scene Generation Using Occlusion-Aware 3D-Conditioned Diffusion} 

\author{Gwanghyun Kim$^{1*}$, Suh Yoon Jeon$^{1*}$, Seunggyu Lee$^{1}$, \stepcounter{footnote} Se Young Chun$^{1,2}$\thanks{} \\
$^1$Dept. of Electrical and Computer Engineering, $^2$INMC \&  IPAI \\
Seoul National University, Republic of Korea \\
{\tt\small \{gwang.kim, euniejeon, leeseunggyu,  sychun\}@snu.ac.kr}
}

\begin{document}

\input{sec_arxiv/0_abstract}
\input{sec_arxiv/1_intro}

\input{sec_arxiv/2_related}

\input{sec_arxiv/3_method}

\input{sec_arxiv/4_experiments}

\input{sec_arxiv/5_conclusions}
{
    \small
    \bibliographystyle{ieeenat_fullname}
    \bibliography{main}
}

\clearpage 
\input{sec_arxiv/supp_arxiv}

\end{document}

%% file: sec_arxiv/0_abstract.tex
\begin{figure*}
\twocolumn[{
\maketitle
\begin{center}
\vspace{-2.em}
\includegraphics[width=0.95\textwidth]{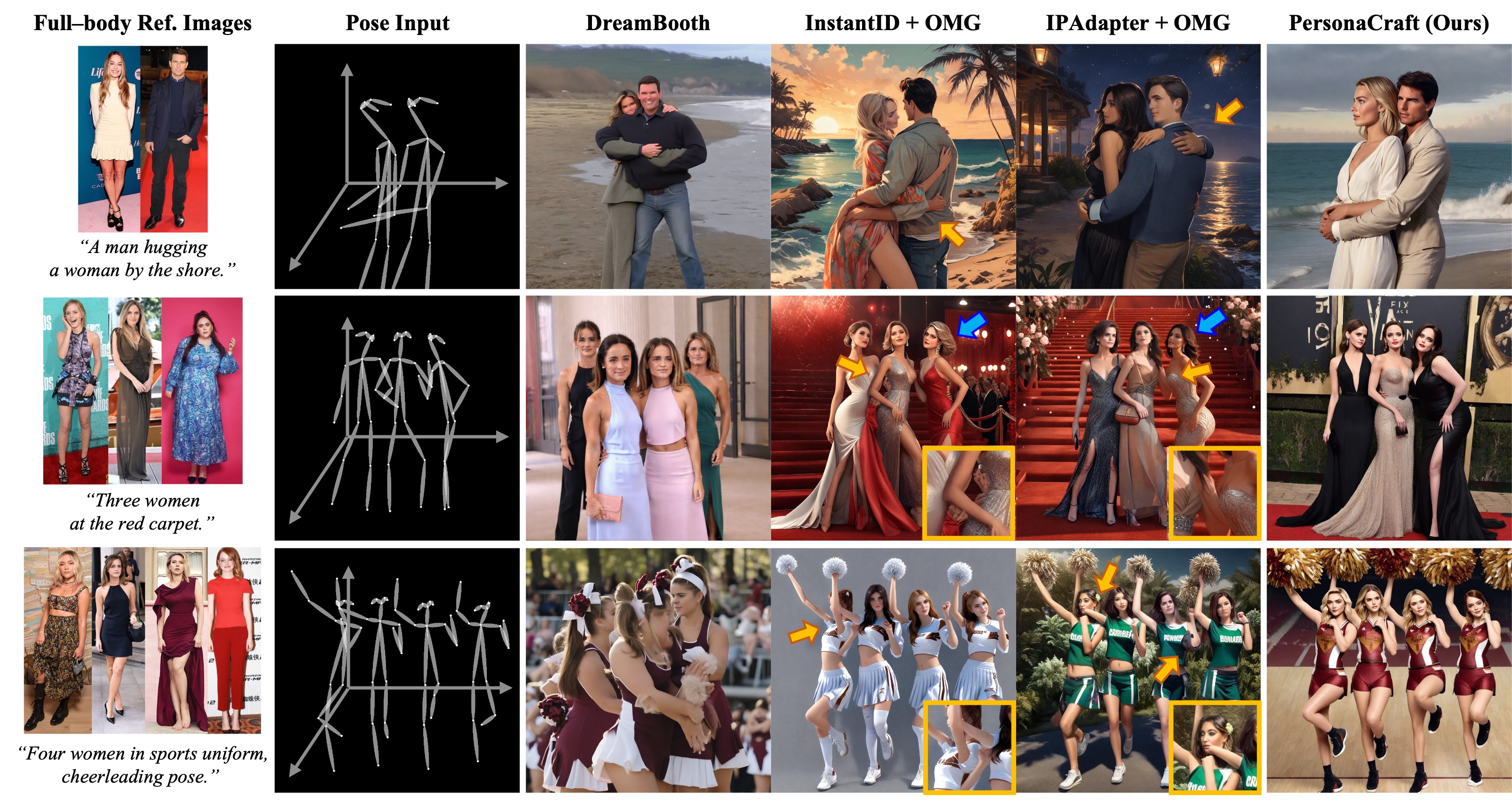}
\vspace{-1.em}
\caption{\textbf{PersonaCraft} generates realistic, personalized images of multiple individuals with complex occlusions, preserving facial identity and body shape using occlusion-aware 3D pose and shape conditioned diffusion. PersonaCraft outperforms baselines in body shape personalization (\textcolor{blue}{blue} arrows indicate failures) and naturalness (\textcolor{yellow}{yellow} arrows highlight artifacts, with zoomed-in views in \textcolor{yellow}{yellow} boxes).}
\label{fig_teaser_1}
\end{center}
}]
\end{figure*}


{
  \renewcommand{\thefootnote}%
    {\fnsymbol{footnote}}
    \footnotetext[1]{Authors contributed equally. \ $^\dagger$Corresponding author.}
  
}

\begin{abstract}
\vspace{-1.5em}


We present PersonaCraft, a framework for controllable and occlusion-robust full-body personalized image synthesis of multiple individuals in complex scenes. Current methods struggle with occlusion-heavy scenarios and complete body personalization, as 2D pose conditioning lacks 3D geometry, often leading to ambiguous occlusions and anatomical distortions, and many approaches focus solely on facial identity. In contrast, our PersonaCraft integrates diffusion models with 3D human modeling, employing SMPLx-ControlNet, to utilize 3D geometry like depth and normal maps for robust 3D-aware pose conditioning and enhanced anatomical coherence. To handle fine-grained occlusions, we propose Occlusion Boundary Enhancer Network that exploits depth edge signals with occlusion-focused training, and Occlusion-Aware Classifier-Free Guidance strategy that selectively reinforces conditioning in occluded regions without affecting unoccluded areas.
PersonaCraft can seamlessly be combined with Face Identity ControlNet, achieving full-body multi-human personalization and thus marking a significant advancement beyond prior approaches that concentrate only on facial identity. Our dual-pathway body shape representation with SMPLx-based shape parameters and textual refinement, enables precise full-body personalization and flexible user-defined body shape adjustments. Extensive quantitative experiments and user studies demonstrate that PersonaCraft significantly outperforms existing methods in generating high-quality, multi-person images with accurate personalization and robust occlusion handling.

\vspace{-2.em}
\end{abstract}

%% file: sec_arxiv/1_intro.tex
\section{Introduction}
Recent advances in personalized image generation~\cite{choi2023custom, hao2023vico, he2023data, ruiz2023dreambooth, ruiz2023hyperdreambooth, smith2023continual, chae2023instructbooth, alaluf2023neural, gal2022image, vinker2023concept, voynov2023p+, pang2023cross, zhang2023compositional, zhao2023catversion, arar2023domain, chen2023subject, gal2023encoder, ma2023unified, shi2023instantbooth, zhou2023enhancing, wei2023elite, gal2023designing, wang2024instantid, li2023photomaker, yan2023facestudio, chen2023anydoor, zhou2023customization, shi2023instantbooth, ye2023ip-adapter} and controllable image generation~\cite{zhang2023adding, liu2023hyperhuman, Guler2018DensePose, zhu2024champ, ju2023humansd, wang2025stable, buchheim2024controlling}, powered by diffusion models~\cite{ho2020denoising, song2020denoising, zhou2023customization, saharia2022photorealistic, betker2023improving, kim2025beyondscene, kim2023datid, kim2023podia, kim2022diffusionclip}, have paved the way for synthesizing multi-human scenes with explicitly defined poses. Despite these breakthroughs, generating images with multiple individuals in complex, occlusion-heavy environments remains a formidable challenge. Conventional approaches that rely on 2D skeleton-based conditioning lack the depth cues necessary to capture overlapping body parts and intricate inter-person interactions, often resulting in ambiguous occlusions and anatomical distortions. Although recent works~\cite{zhu2024champ, buchheim2024controlling, men2024mimo} have begun exploring the use of 3D pose information, they are limited to single-person scenarios, thereby restricting their applicability to multi-person scenes.

Furthermore, achieving full-body personalization remains a significant challenge. Most existing methods prioritize facial identity, often overlooking the distinct characteristics of body shape. Although several approaches~\cite{ye2023ip-adapter, zhou2024storymaker} have enabled full-body personalization using general image encoders~\cite{radford2021learning}, they often struggle to disentangle intrinsic body shape from clothing attributes.

\begin{figure}[!t]
    \centering
    \includegraphics[width=0.5\textwidth]{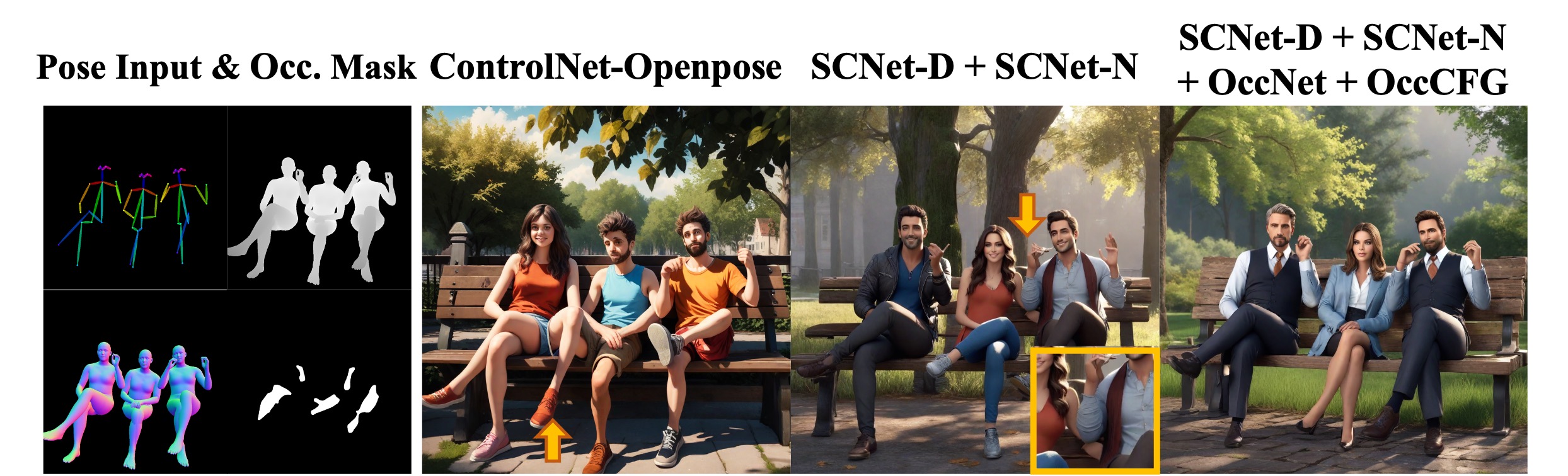}
    \vspace{-2em}
\caption{SMPLx-ControlNet (SCNet) with SMPLx depth and normal improves occlusion handling in pose generation compared to 2D OpenPose, though fine-grained occlusions remain challenging. Using our occlusion-focused methods, OccNet and OccCFG, we achieve superior pose consistency and anatomical coherence.}
    \vspace{-1.8em}
    \label{fig_teaser_2}
\end{figure}

To address these challenges, we propose PersonaCraft, a framework for controllable and occlusion-robust full-body personalized image synthesis of multiple individuals in complex scenes. Firstly, our approach integrates diffusion models with 3D human modeling, employing SMPLx-ControlNet (SCNet), to utilize critical 3D geometric information to overcome the limitations of conventional 2D pose conditioning. Through careful exploration of 3D pose representations, we found that combining SMPLx depth and normal maps offers the most robust conditioning, ensuring high pose consistency and anatomical naturalness. However, despite the benefits of 3D-aware conditioning, handling complex, fine-grained occlusions still remains challenging, as shown in Fig.~\ref{fig_teaser_2}. Secondly, we propose the Occlusion Boundary Enhancer Network (OccNet), which explicitly enhances occluded regions at a finer level  by leveraging depth edge signals that highlight occlusion boundaries. 
Complemented by our Occlusion-Aware Classifier-Free Guidance (OccCFG) strategy, our OccNet adaptively reinforces conditioning in occlusion-heavy areas without compromising contrast in unoccluded regions. 
Together, these components ensure robust anatomical coherence, even in complex multi-human interactions with significant occlusions, as presented in Fig.~\ref{fig_teaser_2}.

Another key advantage of our 3D-aware pose conditioning, beyond enhancing pose accuracy with improved occlusion handling and anatomical coherence, is its ability to encode body shape through SMPLx body shape coefficients. 
Our method can seamlessly be combined with Face Identity ControlNet~\cite{wang2024instantid} to achieve \textit{full-body multi-human personalization}, representing a significant advancement beyond prior methods focused solely on facial identity~\cite{kong2025omg, zhou2024storymaker, he2024uniportrait, zhang2024id, xiao2023fastcomposer, wang2024instantid, ye2023ip-adapter}. Moreover, to overcome the limitations of SMPLx-based body representation, especially in challenging cases where body shape characteristics conflict, we introduce a dual-pathway approach where an optional textual pathway refines body shape. PersonaCraft enables fine-scale, user-defined body shape control through reference-based and interpolation/extrapolation-based adjustments.

Through extensive quantitative and qualitative evaluations, including user studies, our PersonaCraft shows significant improvements in occlusion handling, pose consistency, setting a new benchmark for anatomically precise controllable and personalized multi-human image synthesis.

%% file: sec_arxiv/2_related.tex
\begin{figure*}[!t]
    \centering
    \vspace{-1em}
    \includegraphics[width=0.9\textwidth]{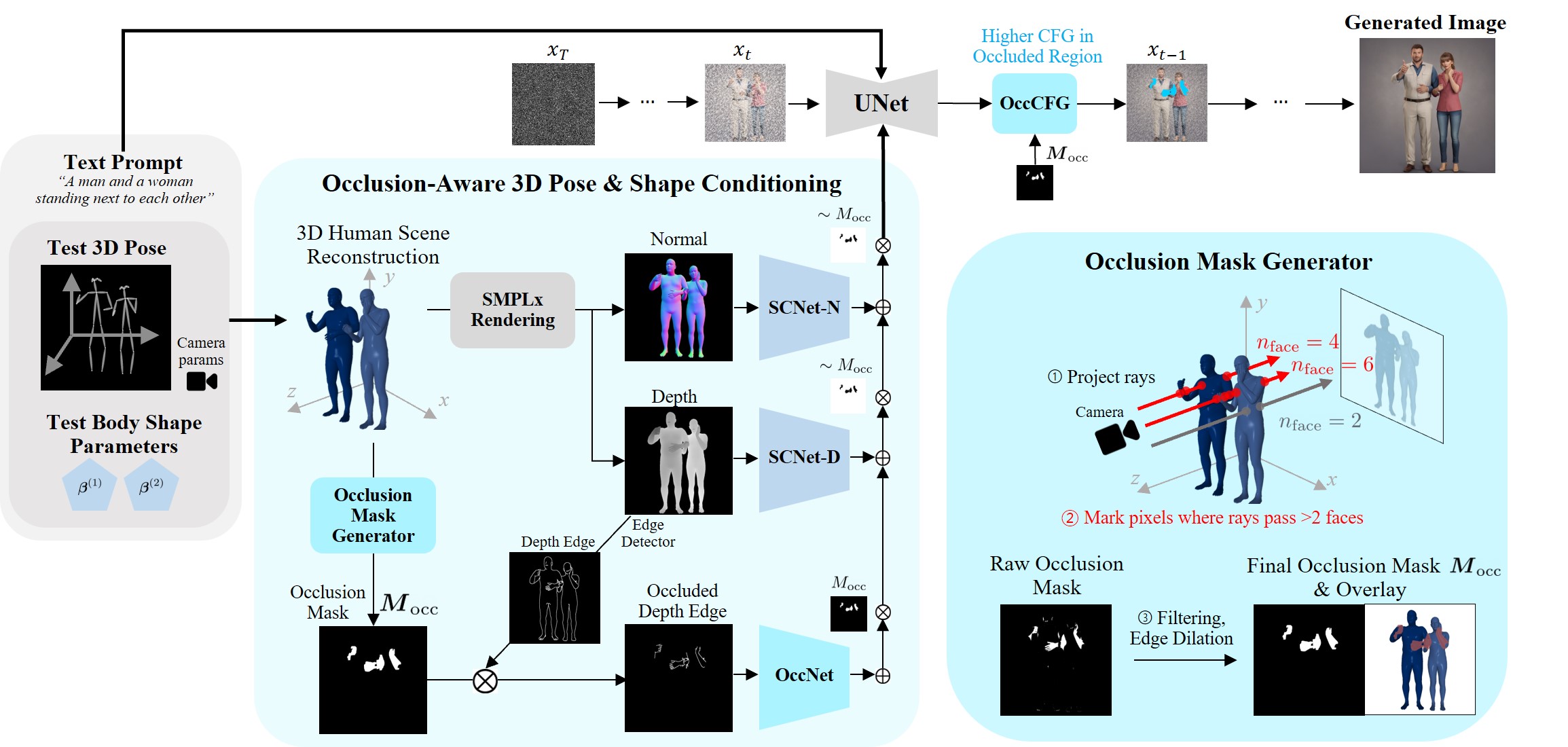}
    \vspace{-0.5em}
\caption{Overview of our occlusion-aware 3D pose and shape conditioning.
We generate SMPLx renderings for SMPLx-ControlNet (\textit{SCNet}) and derive occlusion masks for the Occlusion Boundary Enhancer Network (\textit{OccNet}) by counting intersected faces and masking depth edges. Combining SCNet and OccNet residuals with occlusion masks enables the base U-Net to handle occlusion-aware 3D pose and shape conditioning. Occlusion-aware classifier-free guidance (\textit{OccCFG}) further improves anatomical coherence in occluded regions.}
    \vspace{-1.5em}
    \label{fig_method_1}
\end{figure*}

\section{Related Works}
\vspace{-.5em}

\noindent \textbf{Single-Concept Personalization.} 
Text-to-image (T2I) diffusion models and applications~\cite{ho2020denoising, song2020denoising, zhou2023customization, saharia2022photorealistic, betker2023improving, kim2025beyondscene, kim2023datid, kim2023podia, kim2022diffusionclip} enable single-concept personalization, adapting pre-trained models for individual subjects. Early methods used optimization-based techniques~\cite{choi2023custom, hao2023vico, he2023data, ruiz2023dreambooth, ruiz2023hyperdreambooth, smith2023continual, chae2023instructbooth} or textual embeddings~\cite{alaluf2023neural, gal2022image, vinker2023concept, voynov2023p+, pang2023cross, zhang2023compositional, zhao2023catversion}. LoRA methods~\cite{hu2021lora, tewel2023key} reduced the need for many trainable parameters. Recent works~\cite{arar2023domain, chen2023subject, gal2023encoder, ma2023unified, shi2023instantbooth, zhou2023enhancing, wei2023elite, gal2023designing, wang2024instantid, li2023photomaker, yan2023facestudio, chen2023anydoor, zhou2023customization, shi2023instantbooth, ye2023ip-adapter} like IP-Adapter~\cite{ye2023ip-adapter} and InstantID~\cite{wang2024instantid} adopt modular designs for fast personalization from a single reference image and human-centric modules. However, these methods face challenges in handling complex human poses, neglect body shape preservation, or fail to disentangle identity from clothing in personalized human image synthesis.

\noindent \textbf{Multi-Concept Personalization.}  
Recent multi-concept personalization methods \cite{avrahami2023break, han2023svdiff, liu2023cones, gong2023talecrafter} use cross-attention to prevent concept entanglement. Custom Diffusion \cite{kumari2023multi} integrates models via joint training or constrained optimization. Mix-of-Show \cite{gu2023mix} employs gradient fusion for identity preservation and regional sampling for attribute binding. Modular Customization \cite{po2023orthogonal} isolates concepts via orthogonal directions. FastComposer \cite{xiao2023fastcomposer} enhances training-free personalization. OMG~\cite{kong2025omg} propose 2-stage multi-concept personalization without training. StoryMaker~\cite{zhou2024storymaker} ensures multi-character consistency via segmentation-constrained cross-attention. UniPortrait~\cite{he2024uniportrait} propose ID embeddings and routing modules for high fidelity and editability. ID-Patch~\cite{zhang2024id} mitigates ID leakage and ensures precise identity-position association introducing ID patches. However, existing methods still exhibit limitations in maintaining occlusion robustness and precise control over body shapes.

\noindent \textbf{Controllable Human Generation.} 
Recent works, such as ControlNet~\cite{zhang2023adding} and HyperHuman~\cite{liu2023hyperhuman}, integrate T2I diffusion with 2D skeletons for controllable human synthesis. HumanSD~\cite{ju2023humansd} enhances skeleton-based generation with heatmap-guided denoising, while Stable-Pose~\cite{wang2025stable} improves pose alignment using attention masking. DWpose~\cite{yang2023effective} refines OpenPose~\cite{cao2017realtime} for better skeleton accuracy, and DensePose~\cite{Guler2018DensePose} maps RGB images to 3D surfaces. However, lacking depth cues, these methods struggle with occlusions in multi-human or complex pose scenarios. 
PODIA-3D~\cite{kim2023podia} incorporates depth for pose guidance but not image synthesis. The SMPL~\cite{SMPL:2015} (Skinned Multi-Person Linear Model) and SMPLx~\cite{SMPL-X:2019} are parametric human body models that provide realistic 3D representations of human pose and shape. SMPL models the body using shape parameters and pose parameters with a differentiable blend skinning function, making it widely used in vision and graphics applications. SMPLx extends this by incorporating expressive facial and hand articulation, enabling more detailed human motion modeling. Recent works leverage SMPL for generative tasks: \textbf{Champ}~\cite{zhu2024champ} applies SMPL for single-human video synthesis, while \textbf{HumanLDM}~\cite{buchheim2024controlling} integrates SMPL with \textbf{ControlNet} for stable pose-conditioned human generation. However, these methods focus solely on single-human cases and do not address complex occlusions in multi-human scenarios.

%% file: sec_arxiv/3_method.tex
\section{Proposed Method: PersonaCraft}
We propose PersonaCraft, a framework for controllable and occlusion-robust full-body personalized image synthesis of multiple individuals in complex scenes. PersonaCraft integrates diffusion models with 3D human modeling, leveraging SMPLx-ControlNet (\textit{SCNet}) to incorporate 3D geometry, including depth and normal maps, for robust 3D-aware pose conditioning and enhanced anatomical coherence (Sec.~\ref{sec_method_1}).  To handle fine-grained occlusions, we introduce the Occlusion Boundary Enhancer Network (\textit{OccNet}), which exploits depth edge signals through occlusion-focused training, and the Occlusion-Aware Classifier-Free Guidance (\textit{OccCFG}) strategy, which selectively reinforces conditioning in occluded regions without affecting unoccluded areas (Sec.~\ref{sec_method_2}).  
Furthermore, PersonaCraft seamlessly integrates with Face Identity ControlNet, enabling full-body multi-human personalization and surpassing prior approaches that focus solely on facial identity. Our dual-pathway body shape representation, combining SMPLx-based shape parameters with textual refinement, ensures precise full-body personalization and flexible user-defined body shape adjustments (Sec.~\ref{sec_method_3}).  

Existing methods rely on 2D skeleton-based ControlNets~\cite{zhang2023adding}, which lack depth cues and fail to capture identity-specific body shapes in multi-human scenes. We propose a 3D-aware pose conditioning technique, SMPLx-ControlNet (\textit{SCNet}), that leverages SMPLx~\cite{SMPL-X:2019} to incorporate depth information, enabling occlusion-aware conditioning and body shape control.  

\begin{figure}[!t]
    \centering
    \includegraphics[width=0.45\textwidth]{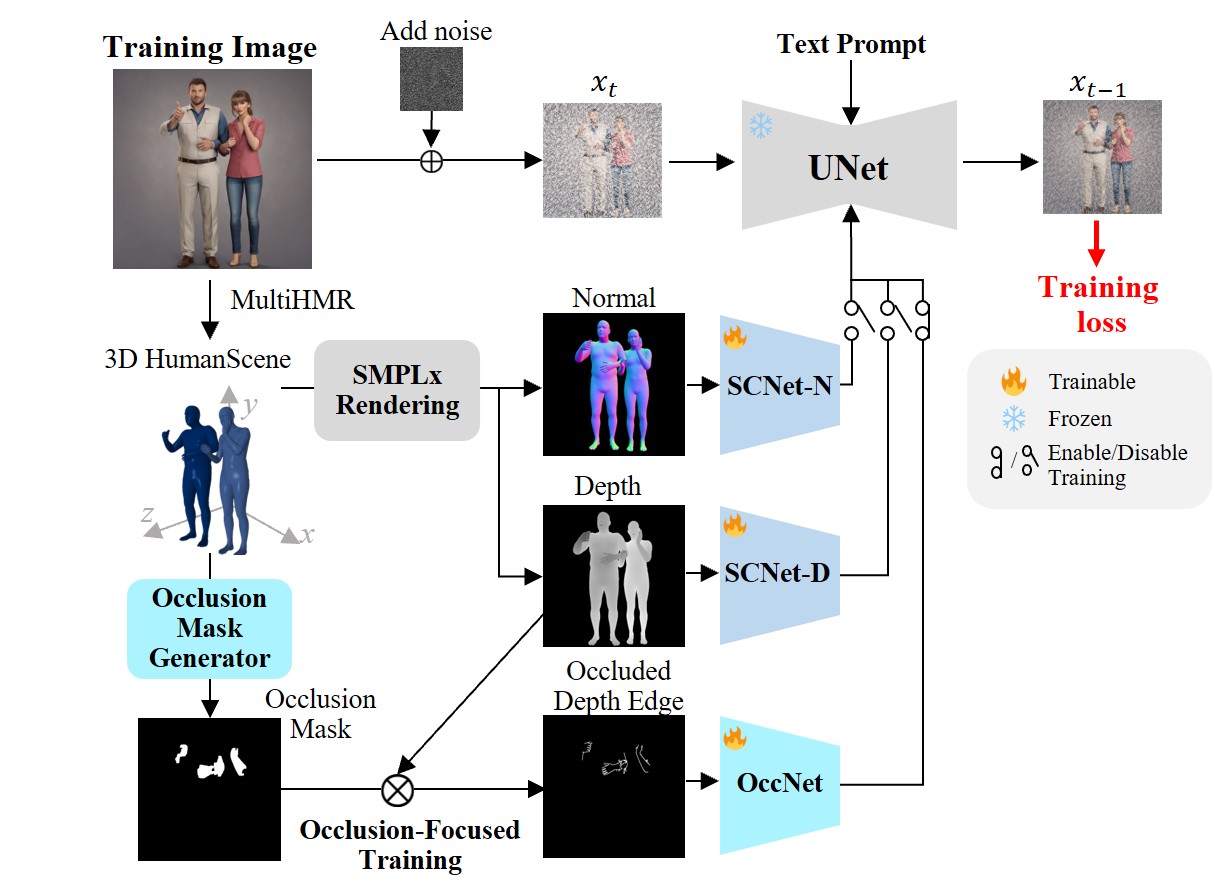}
    \vspace{-1.2em}
    \caption{Training of SMPLx-ControlNet ({SCNet}) and Occlusion Boundary Enhancer Network ({OccNet}). The networks are trained separately, with SMPLx depth, normal maps, and occlusion masks extracted from training images. The pretrained ControlNet~\cite{zhang2023adding} is fine-tuned with these 3D pose representations.} 
    \vspace{-1.3em}
    \label{fig_method_2}
\end{figure}

\subsection{3D-Aware Pose and Shape Conditioning for Multi-Human Generation}
\label{sec_method_1}
Existing methods rely on 2D skeleton-based ControlNets~\cite{zhang2023adding}, which lack depth cues and fail to capture identity-specific body shapes in multi-human scenes. We propose a 3D-aware pose conditioning technique using SMPLx-ControlNet (\textit{SCNet}). By leveraging SMPLx~\cite{SMPL-X:2019}, we represent pose with depth information, enabling occlusion and controlling body shape. As shown in Fig.~\ref{fig_method_1}, given 3D poses $ \{\bm{p}^{(i)}\}_{i=1}^{N_{\text{human}}} $, body shape parameters $ \{\bm{\beta}^{(i)}\}_{i=1}^{N_{\text{human}}} $, and camera parameters $\bm{c}$, we construct a 3D human scene of $N_{\text{human}}$ individuals with occlusions using SMPLx meshes. SMPLx renderings, like depth $\bm{d}_{\text{SMPLx}}$, serve as conditioning signals for the diffusion model, enabling high-fidelity image generation with improved identity preservation and occlusion handling.

As presented in Fig.~\ref{fig_method_2}, we train SCNet $\mathcal{E}^{\text{SC}}_\phi$ by leveraging the pretrained ControlNet~\cite{zhang2023adding} framework, with the following objective:
\begin{equation}
\small
\mathbb{E}_{\bm{x}_0, t, y, \bm{d}_{\text{SMPLx}}, \bm{\epsilon}} \Big[ \Big\Vert \bm{\epsilon} - \bm{\epsilon}_\theta\Big(\bm{x}_t, t, y, \mathcal{E}^{\text{SC}}_\phi(\bm{d}_{\text{SMPLx}})\Big) \Big\Vert_2^2 \Big],
\end{equation}
\normalsize
where $\bm{\epsilon} \sim \mathcal{N}(0,1)$, $\bm{\epsilon}_\theta$ is a base text-to-image diffusion model, \( t \) is the diffusion timesteps \( y \) is the text prompt, \( \bm{x}_t \) is the noisy latent representation of the image at timestep \( t \), and it gets progressively refined during the denoising process. We evaluate different SMPLx renderings—depth, normal, and RGB—for conditioning and find that combining depth SCNet (SCNet-D, \( \mathcal{E}^{\text{SC}}_{\phi_D} \), for occlusion cues) and normal SCNet (SCNet-N, \( \mathcal{E}^{\text{SC}}_{\phi_N} \), for surface orientation) by integrating the residuals from each SCNet provides the most robust pose consistency and anatomical coherence (see Tab.~\ref{tab:ablation_study}). Therefore, we adopt depth-normal as the default 3D representation for SCNet.

\begin{figure}[!t]
    \centering
    \includegraphics[width=0.5\textwidth]{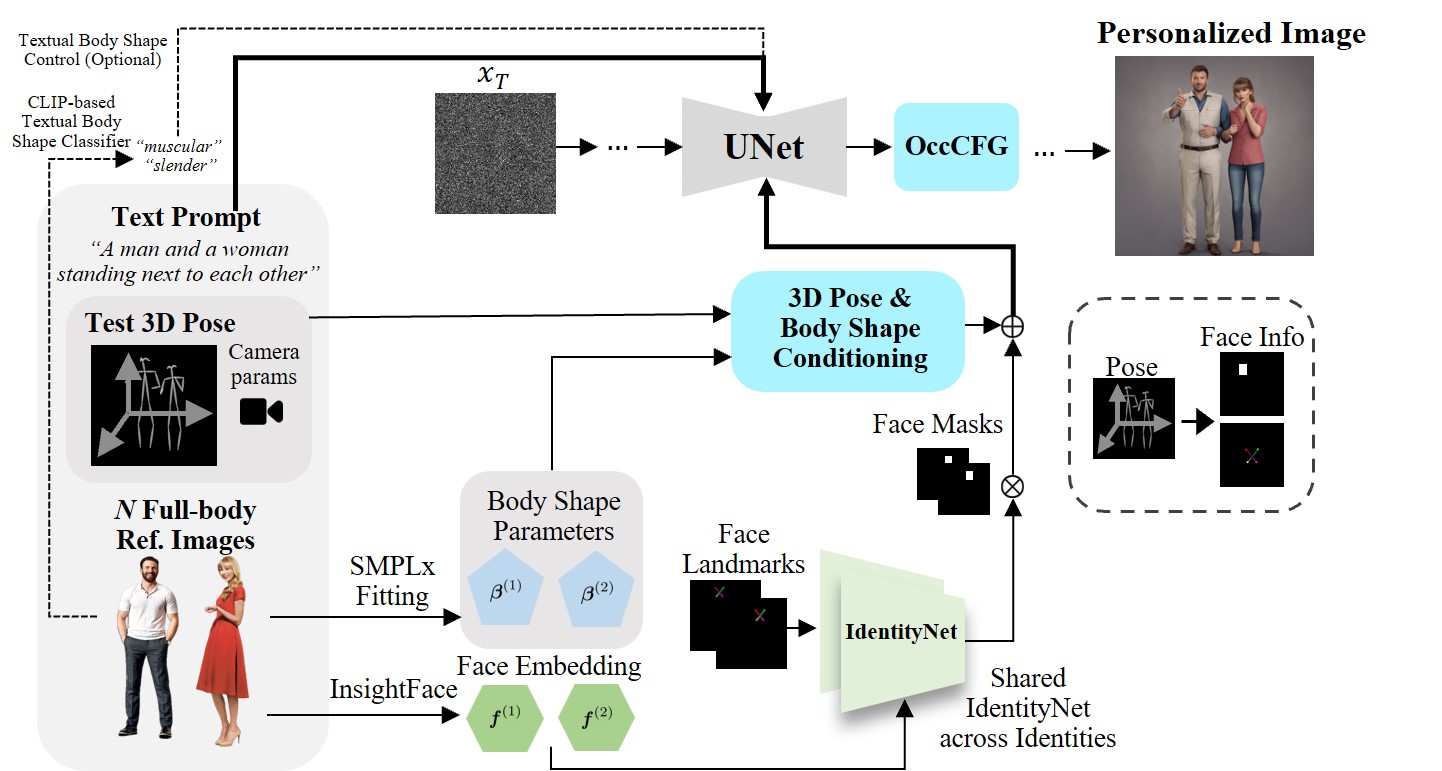}
    \vspace{-1.5em}
\caption{Full pipeline of PersonaCraft for multi-human full-body personalization. By integrating our method with face embeddings from InsightFace~\cite{insightface} and Face Identity ControlNet~\cite{wang2024instantid}, we advance toward full-body multi-human personalization.}
    \vspace{-1.5em}
    \label{fig_method_3}
\end{figure}

\subsection{Occlusion-Focused Generation}
\label{sec_method_2}
Although 3D-aware pose conditioning improves handling occluded regions, generating small, detailed occlusions remains difficult due to inter-person contact and self-occlusion, which cause anatomical distortions. To address this, we introduce occlusion masks from SMPLx-rendered 3D scenes, marking pixels as occluded if covered by multiple surfaces, as shown in Fig.~\ref{fig_method_1} (right). For each pixel $(i,j)$, we count the number of faces $n_{\text{faces}}(i,j)$ that a projected ray passes through. A pixel is considered occluded if the ray intersects more than two surfaces. The occlusion mask \(M_{\text{occ}}\) is defined as:
\begin{equation}
\footnotesize
M_{\text{occ}}(i,j) = 
\begin{cases} 
1, & \text{if } n_{\text{faces}}(i,j) > 2 \\
0, & \text{otherwise}
\end{cases}
\end{equation}
To reduce subtle occlusions and prevent overfitting, we filter small occlusions and refine the boundaries by dilating the mask's edges. This improves the representation of occluded regions, especially in poses with inter-person contact and self-occlusion.

\begin{figure*}[!t]
    \centering
    \vspace{-0.8em}
    \includegraphics[width=0.95\textwidth]{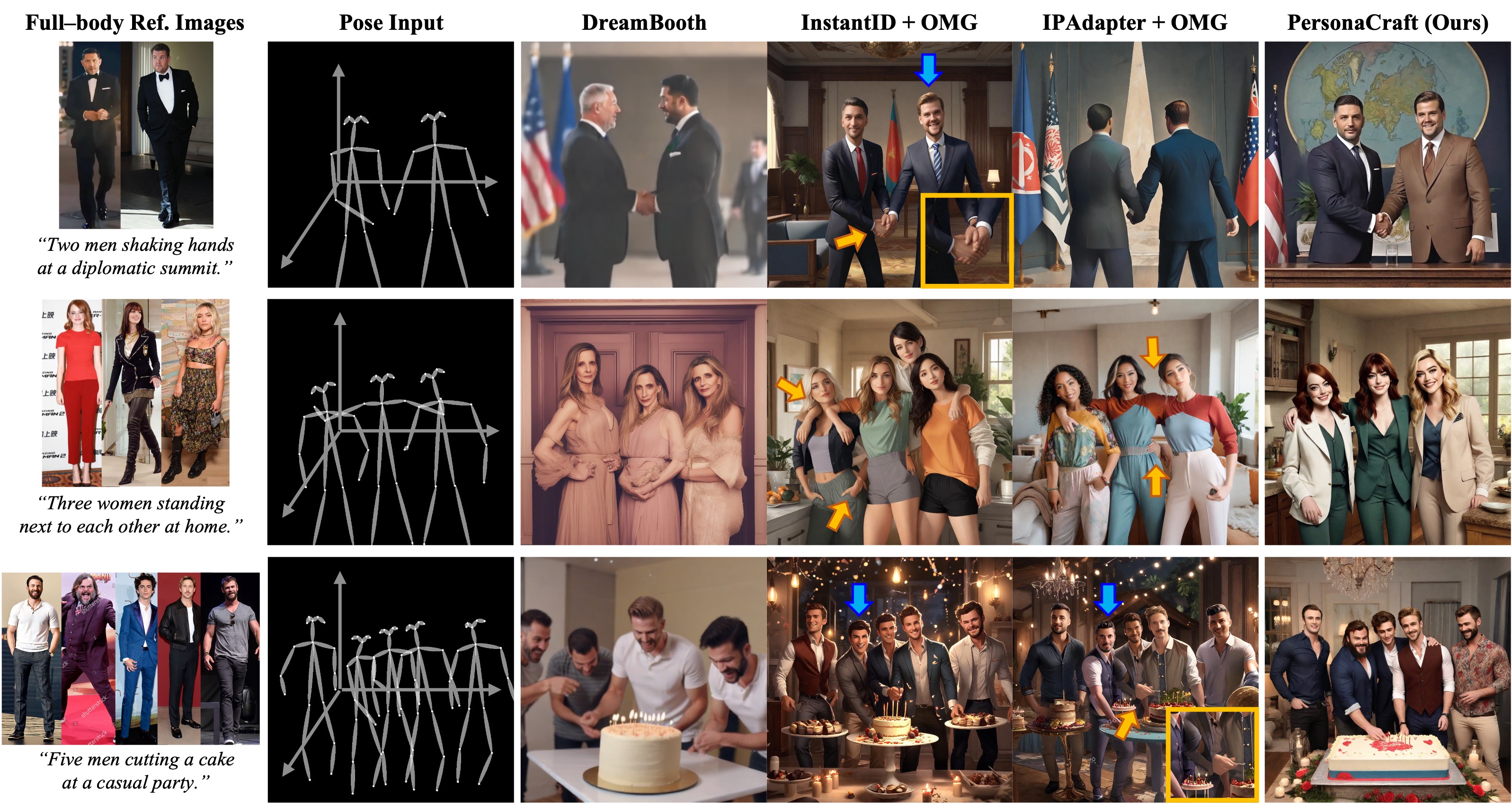}
    \vspace{-0.8em}
    \caption{{Qualitative comparison of personalized multi-human scene generation. \textcolor{blue}{Blue} arrows indicate failures in body shape personalization, while \textcolor{yellow}{yellow} arrows highlight unnatural anatomical structures, with \textcolor{yellow}{yellow} boxes providing zoomed-in views.}}
    \label{fig_results_personalization}
    \vspace{-0.8em}
\end{figure*}


\begin{table*}[!t]\centering
\caption{{Quantitative evaluation of multi-human personalization across face identity, body shape preservation, pose accuracy, text alignment, and image quality. ‘Single’ denotes personalization for a single individual, while ‘Multi’ refers to cases with multiple identities (2–5), with ‘Total’ representing the averaged results.}}
\label{tab:mh_persona}
\scriptsize
\vspace{-1em}
\centering
\begin{adjustbox}{width=0.9\linewidth}
\begin{tabular}{cccccccccccc}
\toprule
\multirow{2}{*}[-0.5em]{\shortstack[c]{Multi-Human\\Personalization}} 
& \multicolumn{3}{c}{Face ID preservation↑} 
& \multicolumn{3}{c}{Body shape preservation↑} 
& \multicolumn{2}{c}{Pose} 
& \multicolumn{1}{c}{Text} 
& \multicolumn{2}{c}{Image quality}\\

\cmidrule(lr){2-4} \cmidrule(lr){5-7} \cmidrule(lr){8-9} \cmidrule(lr){10-10} \cmidrule(lr){11-12}
& Single & Multi & Total & Single & Multi & Total 
& MPJPE (3D)↓ & AP-0.5 (2D)↑ & CLIP sim ↑ & IS ↑ & KID ↓ \\
\midrule
InstantID + OMG & 0.369 & 0.199 & 0.227 & 0.520 & 0.345 & 0.374 & 112.6 & 0.258 & 0.267 & 3.923 & 0.101 \\
IPAdapter+ OMG  & 0.228 & 0.129 & 0.142 & 0.565 & 0.376 & 0.401 & 116.3 & 0.244 & 0.266 & 4.133 & 0.102 \\
IPA-Face + OMG   & 0.310 & 0.166 & 0.188 & 0.515 & 0.341 & 0.367 & 115.4 & 0.249 & 0.267 & 4.021 & 0.102 \\
\textbf{Ours} & \textbf{0.421} & \textbf{0.298} & \textbf{0.317} & \textbf{0.630} & \textbf{0.548} & \textbf{0.560} & \textbf{60.65} & \textbf{0.506} & \textbf{0.273} & \textbf{4.237} & \textbf{0.093} \\
\bottomrule
\end{tabular}
\end{adjustbox}
\vspace{-2em}
\end{table*}

\begin{table}[!t]\centering
\caption{{Top-1 preference rates from the user study on naturalness, face identity, body shape, and text-image correspondence in personalized multi-human scene generation.}}\label{tab:user_study_personalization}
\scriptsize
\vspace{-1em}
\label{tab:user}
\centering
\begin{adjustbox}{width=1\linewidth}
\begin{tabular}{cccccc}\toprule
Top 1  (\%) &Natural.↑ &Face ID↑ &Body shape↑ &Text corr.↑ \\\midrule
Textual Inversion &9.77 &11.46 &10.29 &8.41 \\
DreamBooth &15.4 &11.13 &13.33 &13.01 \\
InstantID + OMG &10.74 &11.65 &11.59 &12.62 \\
IPAdapter + OMG &12.75 &11.78 &12.62 &14.3 \\
IPA-Face + OMG &10.87 &9.51 &10.68 &10.74 \\
\textbf{Ours} &\textbf{40.45} &\textbf{44.47} &\textbf{41.49} &\textbf{40.91} \\
\bottomrule
\end{tabular}
\end{adjustbox}
\vspace{-1.5em}
\end{table}


\noindent \textbf{Occlusion Boundary Enhancer Network.}
Using this occlusion mask, we introduce an Occlusion Boundary Enhancer Network (\textit{OccNet}) by training the model to generate images based solely on \textit{SMPLx depth edges} within the occlusion mask. The SMPLx depth edges, defined as $\bm{e}_{\text{SMPLx}} = \left| \frac{\partial \bm{d}_{\text{SMPLx}}}{\partial x} \right| + \left| \frac{\partial \bm{d}_{\text{SMPLx}}}{\partial y} \right| > \tau,$ where $\tau$ is the edge threshold, highlight occlusion boundaries where depth changes abruptly. Although this information is limited due to the loss of 3D information inside the occluded regions, it serves as a crucial signal by highlighting the occlusion boundaries.

 As shown in Fig.~\ref{fig_method_1}, we provide only the depth edges within the occluded region and train the model to generate images based on this sparse signal, which enables the model to focus more on generating the occluded region while preserving its boundaries, compared to training with the full conditioning. As illustrated in Fig.~\ref{fig_method_2},  we train OccNet $\mathcal{E}^{\text{Occ}}_\omega$ using the following objective:
\begin{equation}
\footnotesize
\mathbb{E}_{\bm{x}_0, t, y, \bm{e}_{\text{SMPLx}}, \bm{M}_{\text{occ}}, \bm{\epsilon}} \left[ \left\Vert \bm{\epsilon} - \bm{\epsilon}_\theta\left(\bm{x}_t, t, y, \mathcal{E}^{\text{Occ}}_\omega\left(\bm{e}_{\text{SMPLx}}\odot \bm{M}_{\text{occ}}\right)\right) \right\Vert_2^2 \right].
\end{equation}

\noindent \textbf{Occlusion-Aware Classifier-Free Guidance.}
We observe that increasing classifier-free guidance (CFG) in occluded regions enhances anatomical consistency by strengthening 3D pose information, resolving local ambiguities. However, a uniformly high CFG strength induce over-saturation in non-occluded areas, as noted in prior works~\cite{sadat2024eliminating, saharia2022photorealistic, kynkaanniemi2025applying}. This highlights the need for a spatially adaptive guidance strategy~\cite{shen2024rethinking} tailored for occlusion in human scene scenarios.  To this end, we introduce Occlusion-Aware Classifier-Free Guidance (\textit{OccCFG}) as shown in Fig.~\ref{fig_method_1}:
\begin{equation}
\footnotesize
    \hat{\bm{\epsilon}} = \bm{\epsilon}_{\text{uncond}} + (k_{\text{occ}} \bm{M}_{\text{occ}} + k_{\text{base}} (1 - \bm{M}_{\text{occ}})) (\bm{\epsilon}_{\text{cond}} - \bm{\epsilon}_{\text{uncond}})
\end{equation}
where \( k_{\text{occ}} \) and \( k_{\text{base}} \) are CFG scales for occluded and non-occluded regions, respectively, \( \bm{M}_{\text{occ}} \) is the occlusion mask, \( \bm{\epsilon}_{\text{uncond}} \) is the unconditional noise prediction, and \( \bm{\epsilon}_{\text{cond}} \) is the conditional noise prediction.



\subsection{Full-Body Personalized Generation}  
\label{sec_method_3}
Another key advantage of SMPLx beyond its robust pose representation is its ability to encode body shape via shape coefficients, enabling a step forward in \textit{full-body multi-human personalization}, in contrast to prior methods that focus solely on facial identity~\cite{kong2025omg, zhou2024storymaker, he2024uniportrait, zhang2024id, xiao2023fastcomposer, wang2024instantid, ye2023ip-adapter}. Furthermore, existing approaches struggle to disentangle garments from body shape, whereas ours explicitly isolates body shape as the true identity-defining feature.


\begin{figure*}[!t]
    \centering
    \vspace{-0.8em}
    \includegraphics[width=0.95\textwidth]{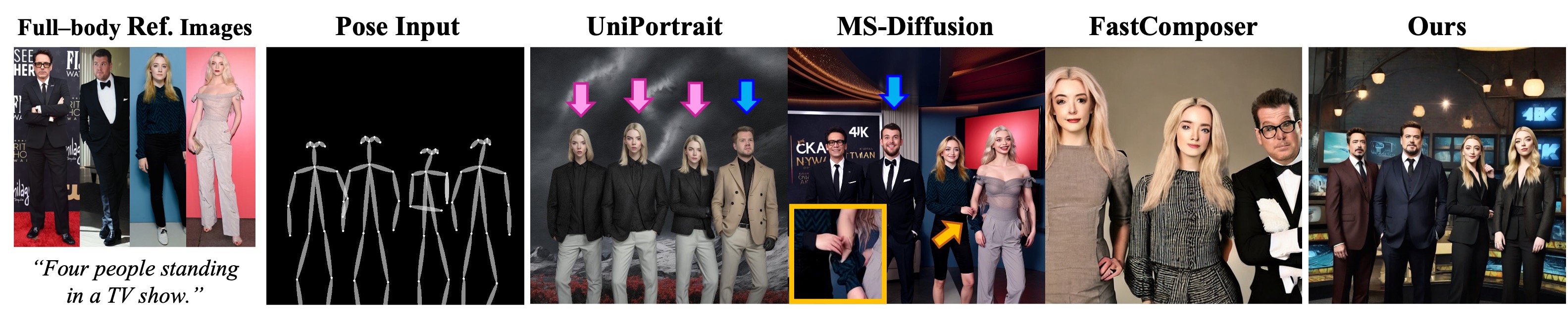}
    \vspace{-.8em}
    \caption{ Comparison of  personalized multi-human scene generation with the methods which are not direct baselines. \textcolor{yellow}{Yellow} arrows highlight anatomical inconsistencies in poses and occlusions, while \textcolor{pink}{Pink} arrows indicate identity mixing or duplication in the baselines. }
    \label{fig_results_personalization_others}
\end{figure*}
\begin{figure*}[!t]
    \centering
    \vspace{-0.8em}
    \includegraphics[width=0.95\textwidth]{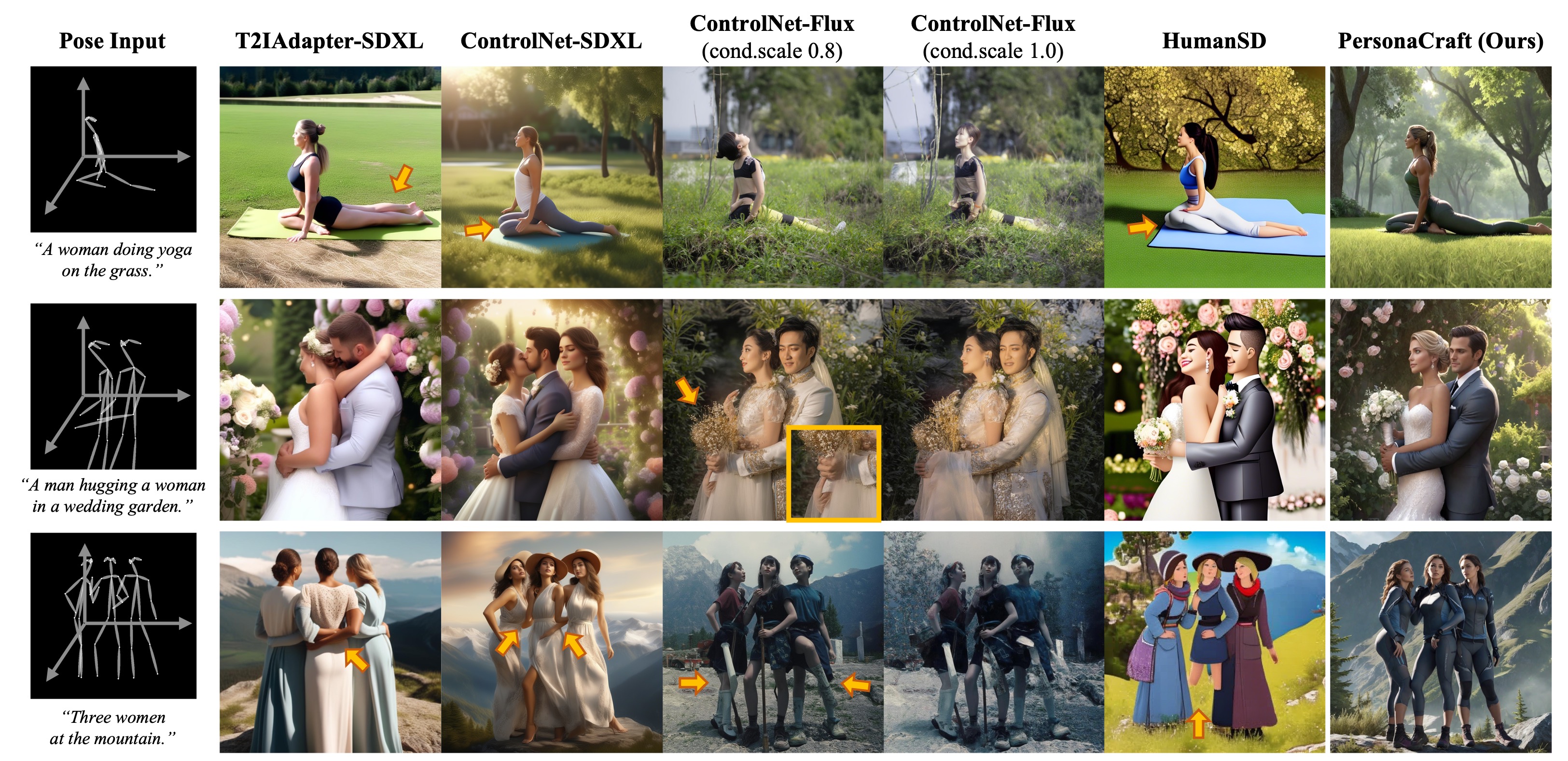}
    \vspace{-1em}
    \caption{Qualitative comparison of pose-controlled multi-human generation.  \textcolor{yellow}{yellow} arrows highlight unnatural anatomical structures, with \textcolor{yellow}{yellow} boxes providing zoomed-in views.}
    \vspace{-1.5em}
    \label{fig_results_pose}
\end{figure*}


\begin{table}[!t]\centering
\caption{{Quantitative evaluation of pose-controlled human generation across pose accuracy, text alignment, and image quality.}}
\label{tab:pose_controlled}
\scriptsize
\vspace{-1em}
\centering
\begin{adjustbox}{width=1\linewidth}
\begin{tabular}{cccccc}
\toprule
\multirow{2}{*}[-0.5em]{\shortstack[c]{Pose-Controlled\\Human Generation}} 
& \multicolumn{2}{c}{Pose} 
& \multicolumn{1}{c}{Text} 
& \multicolumn{2}{c}{Image quality}\\

\cmidrule(lr){2-3} \cmidrule(lr){4-4} \cmidrule(lr){5-6}
& MPJPE (3D) ↓ & AP-0.5 (2D) ↑ & CLIP sim ↑ & IS ↑ & KID ↓ \\
\midrule
T2I Adpater-SDXL & 198.14 &  0.323   & 0.269 & 3.740 & 0.129 \\
ControlNet-SDXL & 114.93 & 0.291 & \textbf{0.285} & 3.062 & 0.138 \\
ControlNet-Flux       & 102.64 & 0.393 & 0.218 & 3.651 & 0.107 \\
HumanSD-SD2      & 103.62 & 0.357 & 0.261 & 3.610 & 0.082 \\
\textbf{Ours}  & \textbf{62.647} & \textbf{0.495} & 0.274 & \textbf{4.113} & \textbf{0.091} \\
\bottomrule
\end{tabular}
\end{adjustbox}
\vspace{-1.5em}
\end{table}


\begin{table}[!t]\centering
\caption{{Top-1 preference rates from the user study on naturalness, pose consistency, and text-image correspondence in pose-controlled multi-human scene generation.}}
\label{tab:user_study_pose_controlled}
\scriptsize
\vspace{-1em}
\centering
\begin{adjustbox}{width=0.8\linewidth}
\begin{tabular}{cccc}\toprule
Top 1 (\%) & Natural.↑ & Pose consistency↑ & Text corr.↑ \\\midrule
T2I Adpater-SDXL & 9.53 & 7.31 & 8.19 \\
ControlNet-SDXL & 17.72 & 18.19 & 19.36 \\
\textbf{Ours} & \textbf{72.75} & \textbf{74.50} & \textbf{72.46} \\
\bottomrule
\end{tabular}
\end{adjustbox}
\vspace{-1.5em}
\end{table}


The overall process of full-body personalized image synthesis is illustrated in Fig.\ref{fig_method_3}. Given full-body reference images $\{I_{\text{ref}}^{(i)}\}_{i=1}^{N_{\text{human}}}$ of $N_{\text{human}}$ individuals, target poses $\{\bm{p}^{(i)}\}_{i=1}^{N_{\text{human}}}$, and a text prompt $y$, our goal is to generate a personalized image $I_P$ that faithfully preserves both face and body identities while ensuring pose consistency. To achieve this, we use MultiHMR~\cite{baradel2025multi} to estimate SMPLx body shape parameters $\bm{\beta}^{(i)}$ and InsightFace~\cite{insightface} to extract face embeddings $\bm{f}^{(i)}$.

We then leverage our depth-normal SCNet $\mathcal{E}^{\text{SC}}_{\phi_D}$-$\mathcal{E}^{\text{SC}}_{\phi_N}$, OccNet $\mathcal{E}^{\text{Occ}}_{\omega}$, and face ControlNet IdentityNet~\cite{wang2024instantidzeroshotidentitypreservinggeneration} $\mathcal{E}^{\text{ID}}_{\psi}$.  
Let $F^{(k)}_\theta(\cdot)$ denote the $k$-th neural block and $\bm{s}^{(k)}$ the corresponding input feature map. 
When $\bm{R}_{\text{sc}} = \mathcal{E}^{\text{SC}}_{\phi_D}(\bm{d}_{\text{SMPLx}}) + \mathcal{E}^{\text{SC}}_{\phi_N}(\bm{n}_{\text{SMPLx}})$,  
$\bm{R}_{\text{occ}} = \mathcal{E}^{\text{Occ}}_{\omega}(\bm{d}_{\text{SMPLx}})$, and  
$\bm{R}^{(i)}_{\text{id}} = \mathcal{E}^{\text{ID}}_{\psi}(\bm{f}^{(i)}, \bm{p}_{\text{face}}^{(i)})$  
are the residual features from each model,
the next feature map $\bm{s}^{(k+1)}$ is obtained by adding these residual features to the neural block output $F^{(k)}_\theta(\bm{s}^{(k)})$, scaled by their respective conditioning weights $\alpha_{\text{sc}}$, $\alpha_{\text{occ}}$, and $\alpha_{\text{id}}$, and modulated by the face masks $\bm{M}_{\text{face}}^{(i)}$ and occlusion mask $\bm{M}_{\text{occ}}$:  
\begin{equation}
\footnotesize
    \begin{split}
        \bm{s}^{(k+1)} &= F^{(k)}_{\theta}(\bm{s}^{(k)}) + \alpha_{\text{sc}}(1 - \bm{M}_{\text{occ}}) \bm{R}_{\text{sc}}  \\
        &\quad + \alpha_{\text{occ}} \bm{M}_{\text{occ}} \bm{R}_{\text{occ}} + \alpha_{\text{id}} \sum_{i=1}^{N_{\text{human}}} \bm{M}_{\text{face}}^{(i)} \ast \bm{R}^{(i)}_{\text{id}}.
    \end{split}
\end{equation}  
\normalsize


\noindent \textbf{Dual-Pathway Body Shape Personalization.}  
We introduce a dual-pathway approach for body shape personalization, leveraging both SMPLx-based and text-based representations.  
Compared to textual descriptions of body shape, SMPLx-based representations provide more spatially fine-grained encoding of the identity. However, they are limited in accurately conveying body composition details, such as fat and muscle distribution~\cite{smpl_faqs}. In challenging cases where body shape characteristics conflict (e.g., a muscular physique with high body fat), our method optionally incorporates a textual pathway to enhance body shape representation. Specifically, a CLIP-based classifier~\cite{radford2021learning} extracts body shape attributes in text, which are applied using regional prompting~\cite{bar2023multidiffusion} and integrated with SCNet. Further details are provided in the supplementary material.


%% file: sec_arxiv/4_experiments.tex
\section{Experiments}
\noindent \textbf{Implementation Details.} PersonaCraft is built on Stable Diffusion XL (SDXL)\cite{podell2023sdxl}. We extend the MPII dataset\cite{andriluka20142d} with textual descriptions~\cite{khan2024focusclipmultimodalsubjectlevelguidance} and 3D human reconstructions from MultiHMR~\cite{baradel2025multi} using SMPLx~\cite{SMPL-X:2019}. SCNet and OccNet were fine-tuned on ControlNet~\cite{zhang2023adding} with a learning rate of 1e-5, Adam optimizer, batch size of 2, for 50,000 iterations on an NVIDIA A100 GPU. We use 30 denoising steps. We set the base CFG scale to 3 (\(k_{\text{base}} = 3\)) and increased the scale for occluded regions to 5 (\(k_{\text{occ}} = 5\)) to enhance occlusion handling. We set all conditioning scales (\(\alpha_{\text{sc}}\), \(\alpha_{\text{occ}}\), and \(\alpha_{\text{id}}\)) to 0.8 across all cases.


\begin{figure}[!t]
    \centering
    \includegraphics[width=0.5\textwidth]{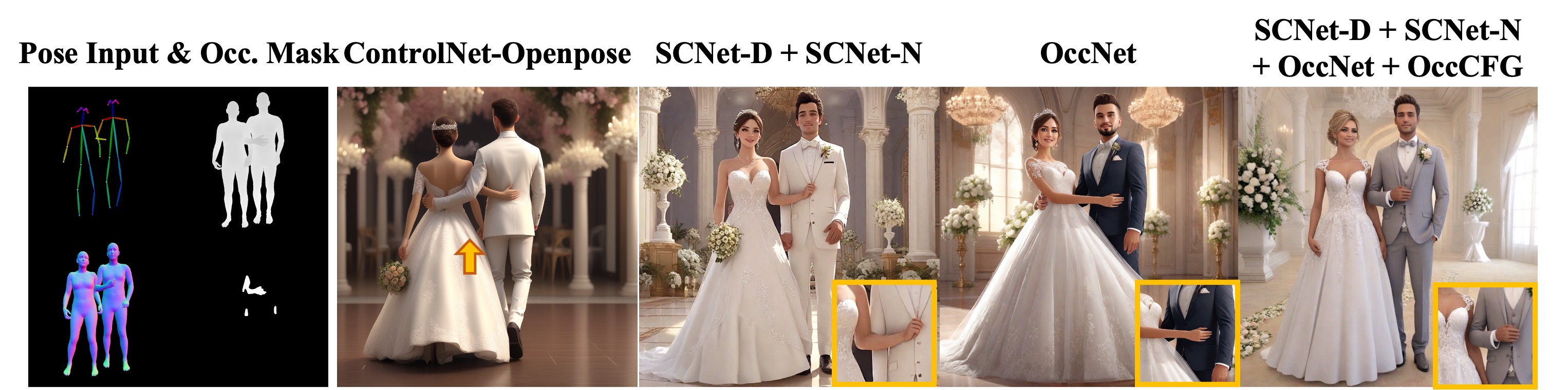}
    \vspace{-2em}
    \caption{Effect of our occlusion-aware 3D pose \& shape conditioning components. D and N denote the conditioning types of SCNet: depth, normal.}
    \vspace{-1.5em}
    \label{fig_pose_analysis2}
\end{figure}

\begin{figure}[!t]
    \centering
    \includegraphics[width=0.5\textwidth]{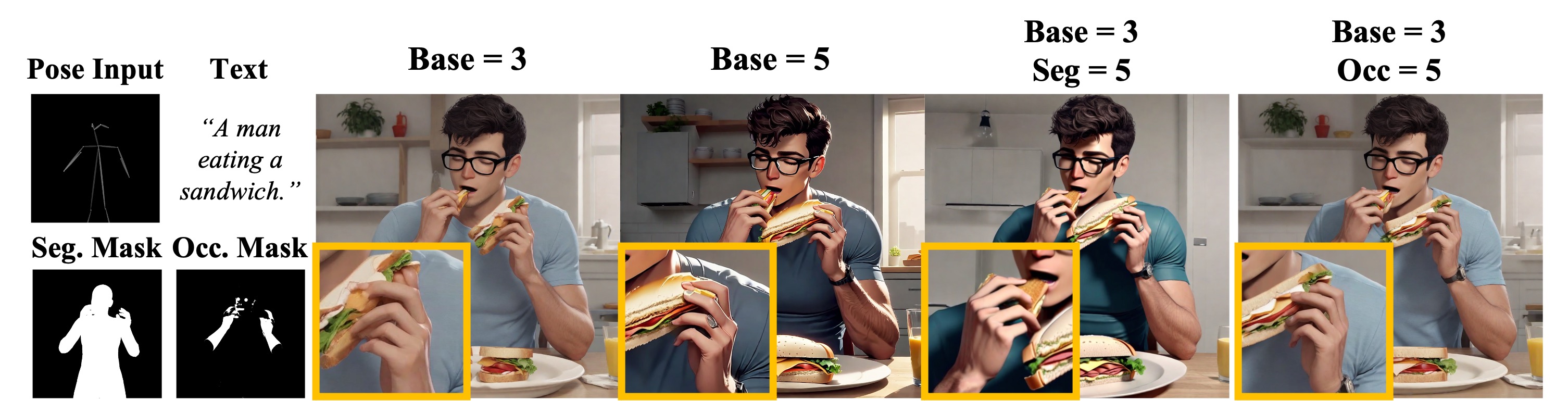}
    \vspace{-2em}
\caption{Effect of our occlusion-aware CFG (OccCFG). Each label refers to the CFG scale applied to specific regions. "Base=3" means the CFG scale is 3 for all regions, while "Base=3, Seg=5" indicates that the CFG scale is 5 for the human segmentation region. "Occ" refers to the occlusion mask region.}
    \label{fig_cfg_analysis}
    \vspace{-0.8em}
\end{figure}

\begin{table}[!t]\centering
\caption{Ablation study on SCNet, OccNet, and OccCFG for pose consistency. D, N, and R denote the conditioning types of SCNet: depth, normal, and RGB representations, respectively.}
\label{tab:ablation_study}
\scriptsize
\vspace{-1em}
\centering
\begin{adjustbox}{width=1\linewidth}
\begin{tabular}{ccc}
\toprule
\multirow{2}{*}[-0.3em]{{Ablation Study}} & \multicolumn{2}{c}{Pose} \\
\cmidrule(lr){2-3}
 & MPJPE (3D)\(\downarrow\) & AP-0.5 (2D)\(\uparrow\) \\
\midrule
SCNet-D                      & 71.20  & 0.423 \\
SCNet-N                        & 79.30  & 0.398 \\
SCNet-R                        & 82.52 & 0.394 \\ 
SCNet-D + SCNet-N                & 63.16   & 0.494 \\ \midrule
SCNet-D + SCNet-N + OccNet        & 62.92  & \textbf{0.499} \\
SCNet-D + SCNet-N + OccNet + OccCFG & \textbf{62.64}  & 0.495 \\
\bottomrule
\end{tabular}
\end{adjustbox}
\vspace{-2em}
\end{table}
\noindent \textbf{Metrics.}
For personalized multi-human scene generation, face identity preservation is measured using FaceNet~\cite{schroff2015facenet}, following FastComposer~\cite{xiao2023fastcomposer}. Body shape preservation is evaluated via cosine similarity of SMPLx body shape parameters $\bm{\beta}$. Pose consistency is assessed using MPJPE (3D)\cite{ionescu2013human3}, while 2D pose estimation is evaluated with AP-0.5 (2D) by comparing the target pose to the estimated pose from generated images. Text-image alignment is measured via CLIP-L/14 similarity. Image quality is quantified using Inception Score (IS)\cite{salimans2016improved} and Kernel Inception Distance (KID)~\cite{binkowski2018demystifying}. A user study evaluates perceptual quality.

For pose-controlled multi-human scene generation, the same key metrics are applied: MPJPE (3D) and AP-0.5 (2D) for pose accuracy, CLIP similarity for text-image alignment, and IS/KID for image quality.

\noindent \textbf{Test Dataset.} We test on the COCO-WholeBody dataset~\cite{jin2020whole}, containing 1$\sim$5 people per image. A total of 1,000 images were selected (200 per group). Text prompts were extracted via BLIP~\cite{li2023blip}. MultiHMR~\cite{multi-hmr2024} provided 2D poses and 3D conditioning.

Additional details on experiments, results (including qualitative results), stylized artwork, PersonaCraft with LoRA and other face control modules, as well as ablation studies, are provided in the supplementary material.

\subsection{Personalized Multi-Human Scene Generation}
\noindent \textbf{Baselines.} We mainly compared PersonaCraft with baseline methods for single-shot, multi-identity, and pose-controllable synthesis, all implemented using SDXL~\cite{podell2023sdxl}. Key baselines include OMG~\cite{kong2025omg} with InstantID/IPAdapter/IPAdapter-Face (IPA-Face)~\cite{wang2024instantid, ye2023ip-adapter}, and 2D pose ControlNet~\cite{zhang2023adding}. We also evaluated optimization-based methods like DreamBooth~\cite{ruiz2023dreambooth} and Texture Inversion~\cite{gal2022image}. Both qualitative and quantitative comparisons, as well as a user study, were conducted. Additionally, qualitative comparisons were made with UniPortrait~\cite{he2024uniportrait}, MS-Diffusion~\cite{wang2024ms}, and FastComposer~\cite{xiao2024fastcomposer}.

\noindent \textbf{Qualitative Results.} As shown in Fig.~\ref{fig_results_personalization}, PersonaCraft surpasses baselines in multi-identity scene generation, particularly in identity preservation and occlusion handling. Our 3D-aware conditioning and occlusion-focussed generation enhance the preservation of body shapes and identity features, and ensures pose fidelity, producing realistic generations with fewer artifacts. Baseline methods struggle with occlusion and body shape accuracy, often leading to distorted identities and unrealistic compositions, especially in multi-subject scenes. PersonaCraft excels in occlusion-aware full-body personalization, maintaining consistent identities across varying poses.

\noindent \textbf{Quantitative Results.} We evaluate PersonaCraft on face ID preservation, body shape, pose accuracy, text-image alignment, and image quality. As shown in Tab.~\ref{tab:mh_persona}, our method consistently outperforms baselines in identity and body shape preservation. It achieves the lowest MPJPE (3D) and highest AP (2D), indicating superior alignment with input poses. Furthermore, PersonaCraft outperforms baselines in IS and KID, producing perceptually aligned generations.

\noindent \textbf{User Study.} We conducted a user study evaluating naturalness, identity, body shape, and text-image correspondence across baselines and our method. With {18,540 responses from 103} participants, Tab.~\ref{tab:user_study_personalization} shows our method achieved the highest Top-1 preference across all metrics, demonstrating superior perceptual quality in personalized multi-human scene generation.

\subsection{Pose-Controlled Multi-Human Scene Generation}
\noindent \textbf{Baselines.} Key baselines include ControlNet-SDXL~\cite{zhang2023adding,podell2023sdxl}, T2I Adapter-SDXL~\cite{mou2023t2i}, ControlNet-Flux~\cite{flux2024, zhang2023adding}, and HumanSD~\cite{flux2024}. Both qualitative and quantitative comparisons were conducted for all baselines. Additionally, a user study was performed specifically for SDXL-based baselines, including ControlNet-SDXL and T2I Adapter-SDXL.


\begin{figure}[!t]
    \centering
    \includegraphics[width=0.5\textwidth]{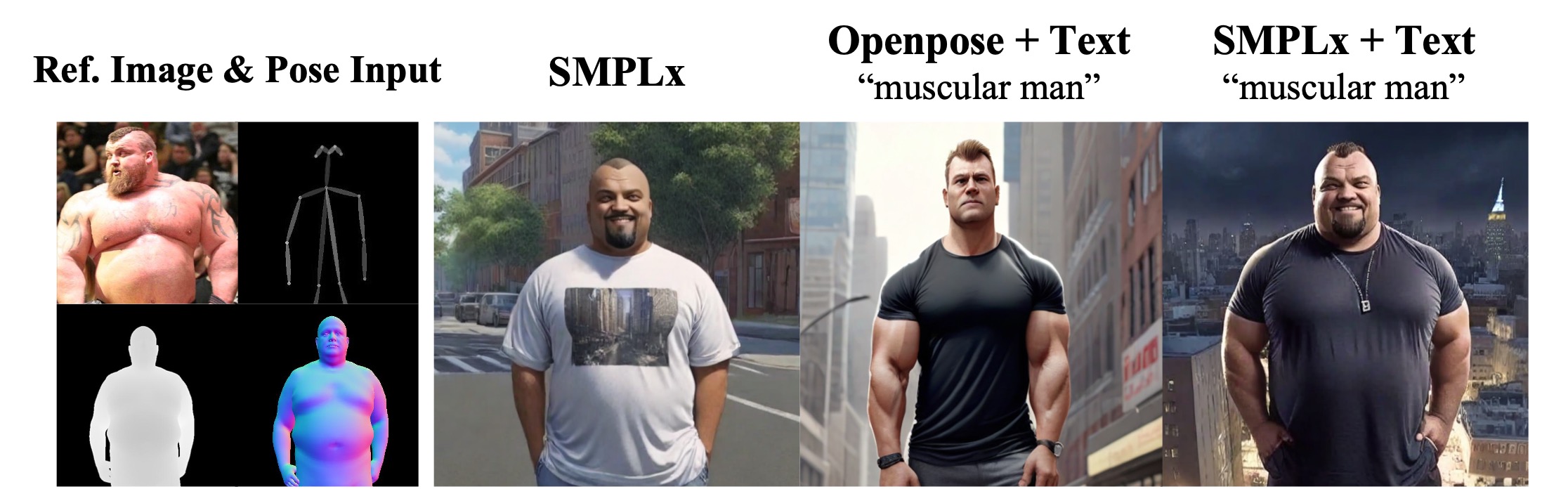}
    \vspace{-2em}
    \caption{Effectiveness of dual-pathway body shape personalization in challenging cases where body shape characteristics conflict (e.g., a muscular physique with high body fat).}
    \label{fig_dual_path}
    \vspace{-1.em}
\end{figure}

\begin{figure}[!t]
    \centering
    \includegraphics[width=0.5\textwidth]{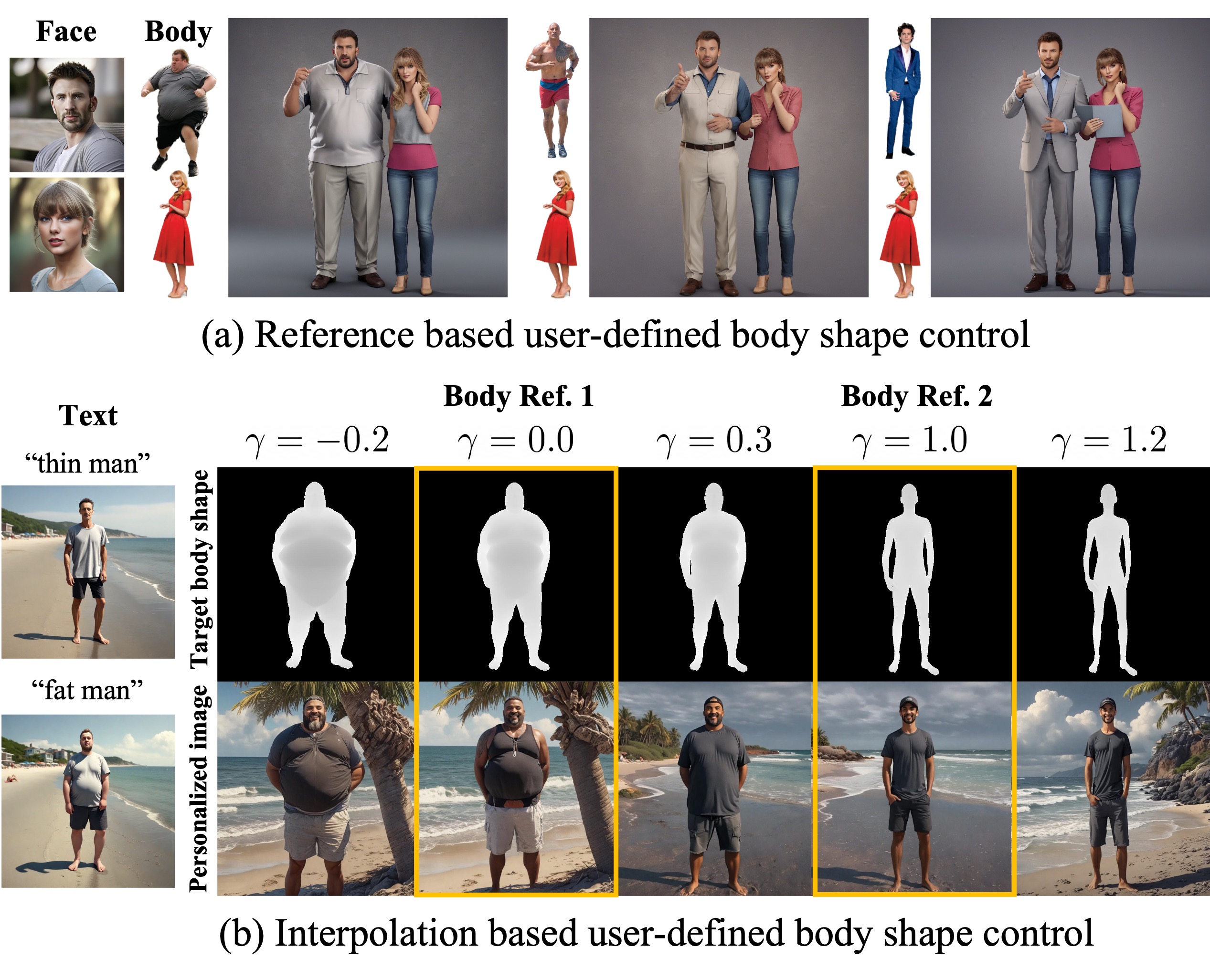}
    \vspace{-2.2em}
    \caption{Result of PersonaCraft's user-defined body shape control. (a) Reference-based body shape control. (b) Interpolation and extrapolation-based control, where $\gamma$ controls the interpolation or extrapolation between the reference body shapes. }
    \vspace{-1.5em}
    \label{fig6}
\end{figure}


\noindent \textbf{Qualitative Results.} We evaluate pose-controlled multi-human scene generation by comparing PersonaCraft with existing methods in both single-human and multi-human settings (Fig.~\ref{fig_results_pose}). Our method achieves superior pose alignment while preserving identity consistency and body shape fidelity. Baselines relying on 2D pose conditioning (e.g., T2IAdapter-SDXL~\cite{mou2023t2i}, ControlNet-SDXL~\cite{zhang2023adding,podell2023sdxl}) struggle with occlusion handling, leading to misaligned poses and distorted structures. ControlNet-Flux~\cite{flux2024,zhang2023adding} exhibits instability: at a lower conditioning scale (0.8), it fails to maintain anatomical coherence, while at a higher scale (1.0), it improves pose accuracy but reduces image fidelity. HumanSD~\cite{ju2023humansd} lacks precise pose control, producing unrealistic compositions. In contrast, PersonaCraft balances pose accuracy, identity preservation, and body realism, demonstrating the efficacy of our 3D-aware pose conditioning.

\noindent \textbf{Quantitative Results.} We evaluate pose-controlled multi-human scene generation in terms of pose accuracy, text-image correspondence, and image quality. As shown in Tab.~\ref{tab:pose_controlled}, PersonaCraft achieves the lowest MPJPE (3D) and highest AP-0.5 (2D), demonstrating precise pose alignment and keypoint localization. Compared to ControlNet-based baselines, which suffer from structural inconsistencies and occlusion issues, our method maintains accurate pose fidelity while preserving identity and body shape. PersonaCraft also achieves the highest IS and lowest KID, indicating superior perceptual realism. While text-image correspondence scores are comparable across methods, prior studies~\cite{ku2023viescore, kim2025beyondscene} highlight CLIP’s limitations in evaluating complex multi-human scenes. A user study further analyzes perceptual quality.

\noindent \textbf{User Study.} We conducted a user study to assess naturalness, pose consistency, and text-image correspondence. Participants ranked their top three preferences among T2I Adapter-SDXL, ControlNet-SDXL, and PersonaCraft, collecting 15,390 responses from 114 participants. As shown in Tab.~\ref{tab:user_study_pose_controlled}, PersonaCraft achieved the highest Top-1 preference across all metrics, excelling in naturalness (72.75\%), pose consistency (74.50\%), and text alignment (72.46\%). These results confirm that our occlusion-aware 3D pose conditioning significantly improves perceptual quality, producing well-posed, visually coherent multi-human scenes.

\subsection{Ablation Studies and Analysis}

\noindent \textbf{Comparison of 3D Pose Representations for SCNet.}
In Tab.~\ref{tab:ablation_study}, we evaluate different SMPLx renderings—depth, normal, and RGB—as conditioning inputs for SCNet. We find that depth yields the best pose consistency among these representations. Additionally, combining the depth and normal representations achieves the best performance. Therefore, we adopt this combination as the base conditioning for SCNet.

\noindent \textbf{Effectiveness of OccNet and OccCFG.}
In Tab.~\ref{tab:ablation_study}, we also evaluate the effectiveness of OccNet and OccCFG. We found that OccNet and OccCFG improves pose consistency.
Also, as analyzed in Fig.~\ref{fig_pose_analysis2}, effect of our Occlusion-aware 3D pose \& shape conditioning components. Using only 2D pose leads to front-back ambiguities and structural inconsistencies, while relying solely on 3D pose (SCNet) struggles with fine-grained occluded regions. Using only OccNet improves occluded region synthesis but fails to maintain overall pose alignment. Our full model (SCNet + OccNet + OccCFG) effectively preserves pose structure while handling occlusions. 
In Fig.~\ref{fig_cfg_analysis}, we present a detailed analysis of OccCFG. We observe that increasing classifier-free guidance (CFG) in occluded regions improves anatomical consistency by leveraging stronger 3D pose information, effectively resolving local ambiguities. However, uniformly high CFG strength leads to over-saturation in non-occluded areas. We found that applying CFG only in the human segmentation region also results in high CFG, whereas our OccCFG avoids issues in unoccluded regions while maintaining effective guidance in occluded areas.

\noindent \textbf{Dual-Pathway Body Personalization.}
As shown in Fig~\ref{fig_dual_path}, the SMPLx-based pathway captures fine-grained identity features but struggles with detailed body composition (e.g., muscle vs. fat distribution). The text-based pathway refines body shape using semantic attributes. Our approach combines both for more accurate and realistic personalization.

\subsection{User-Defined Body Shape Control}
A key feature of PersonaCraft is its user-defined body shape control, as shown in Fig.~\ref{fig6}. Users can select a reference to adapt body characteristics for personalized images. The system also allows body proportions to be adjusted through interpolation or extrapolation between references, offering enhanced customization for precise body shape control.

%% file: sec_arxiv/5_conclusions.tex
\section{Limitations}
While our method is versatile and can be applied to other ControlNet models, the performance of our face personalization depends significantly on the underlying face identity network. Additionally, the accuracy of 3D human model fitting is dependent on the performance of the fitting algorithm used. Variations in the quality of the fitting process may impact the final output, especially in cases where the reference data is incomplete or inaccurate. Failure cases can be found in the supplementary material.

\section{Conclusion}
We present PersonaCraft, a framework for occlusion-robust, full-body personalized image synthesis in complex multi-human scenes. PersonaCraft integrates diffusion models with 3D human modeling, using SMPLx-ControlNet for enhanced pose conditioning.
To address occlusions at a fine level, we introduce Occlusion Boundary Enhancer and Occlusion-Aware Classifier-Free Guidance strategy, improving conditioning in occluded areas. Seamlessly combined with Face Identity ControlNet, our approach enables full-body personalization beyond facial identity. Extensive experiments show PersonaCraft outperforms prior arts in personalization accuracy and occlusion robustness in complex scenes.

%% file: sec_arxiv/supp_arxiv.tex
\appendix

\twocolumn[{%
\renewcommand\twocolumn[1][]{#1}%

\begin{center}
\bigskip 
\bigskip 
\textbf{\Large PersonaCraft: Personalized and Controllable Full-Body Multi-Human Scene Generation Using Occlusion-Aware 3D-Conditioned Diffusion \\ (Supplementary Material) \\}

\bigskip 
\bigskip 
\maketitle
 
\end{center}%
}]

\renewcommand{\theequation}{S\arabic{equation}}
\renewcommand{\thefigure}{S\arabic{figure}}
\renewcommand{\thetable}{S\arabic{table}}

\section{Additional Results}

\subsection{Personalized Multi-Human Scene Generation}
\begin{figure*}[!h]
    \centering
    \includegraphics[width=.75\textwidth]{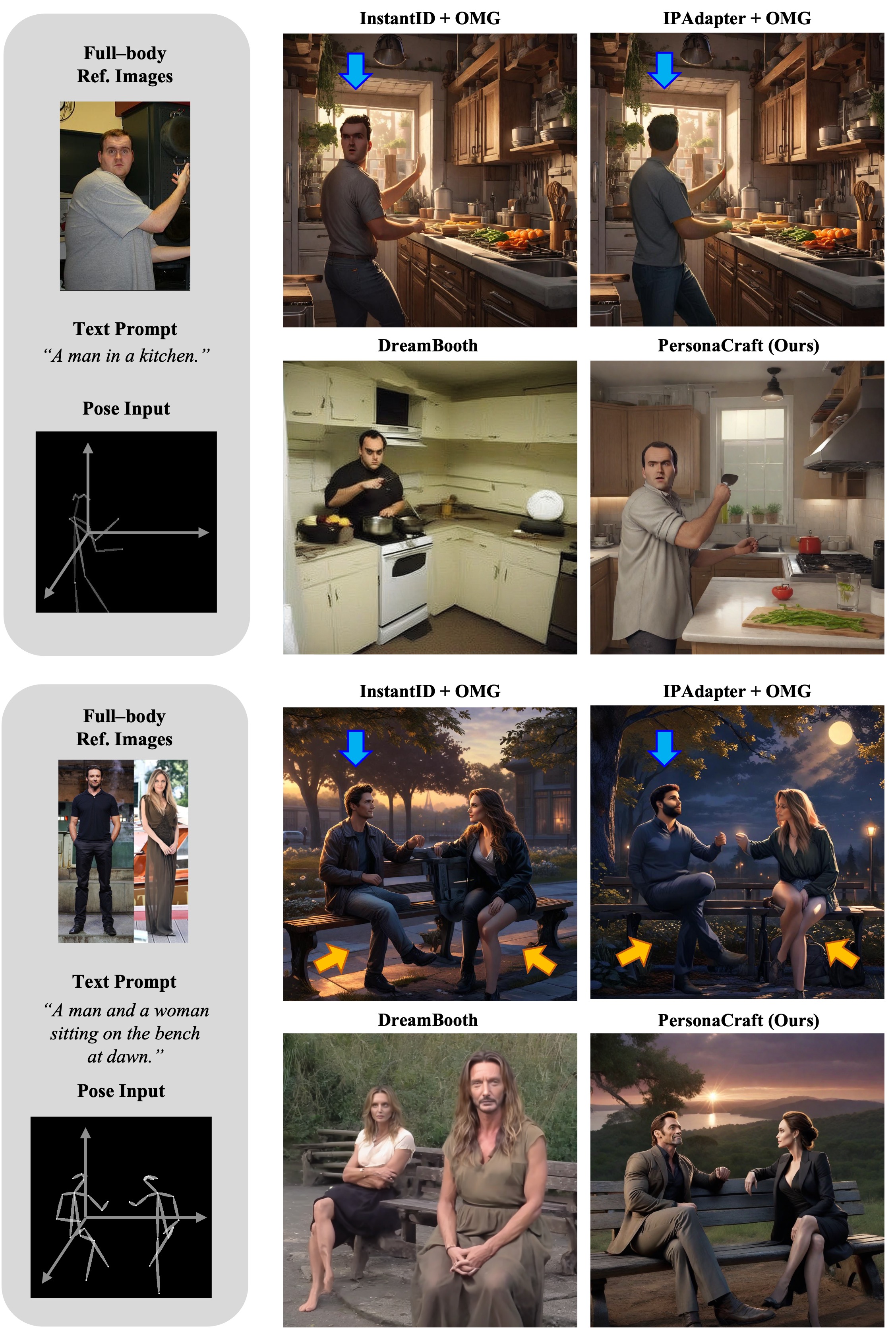}
    \caption{{Additional comparison of generated images. Yellow arrows highlight anatomical issues due to 2D pose limitations. Blue arrows refer to the individuals evaluated for correct body shape preservation. PersonaCraft excels in identity, body shape consistency, and naturalness while being over twice as fast as OMG-based methods~\cite{wang2024instantid, ye2023ip-adapter, kong2025omg}.}}
    \label{fig_supp_more_results1}
\end{figure*}

\begin{figure*}[!h]
    \centering
    \includegraphics[width=.7\textwidth]{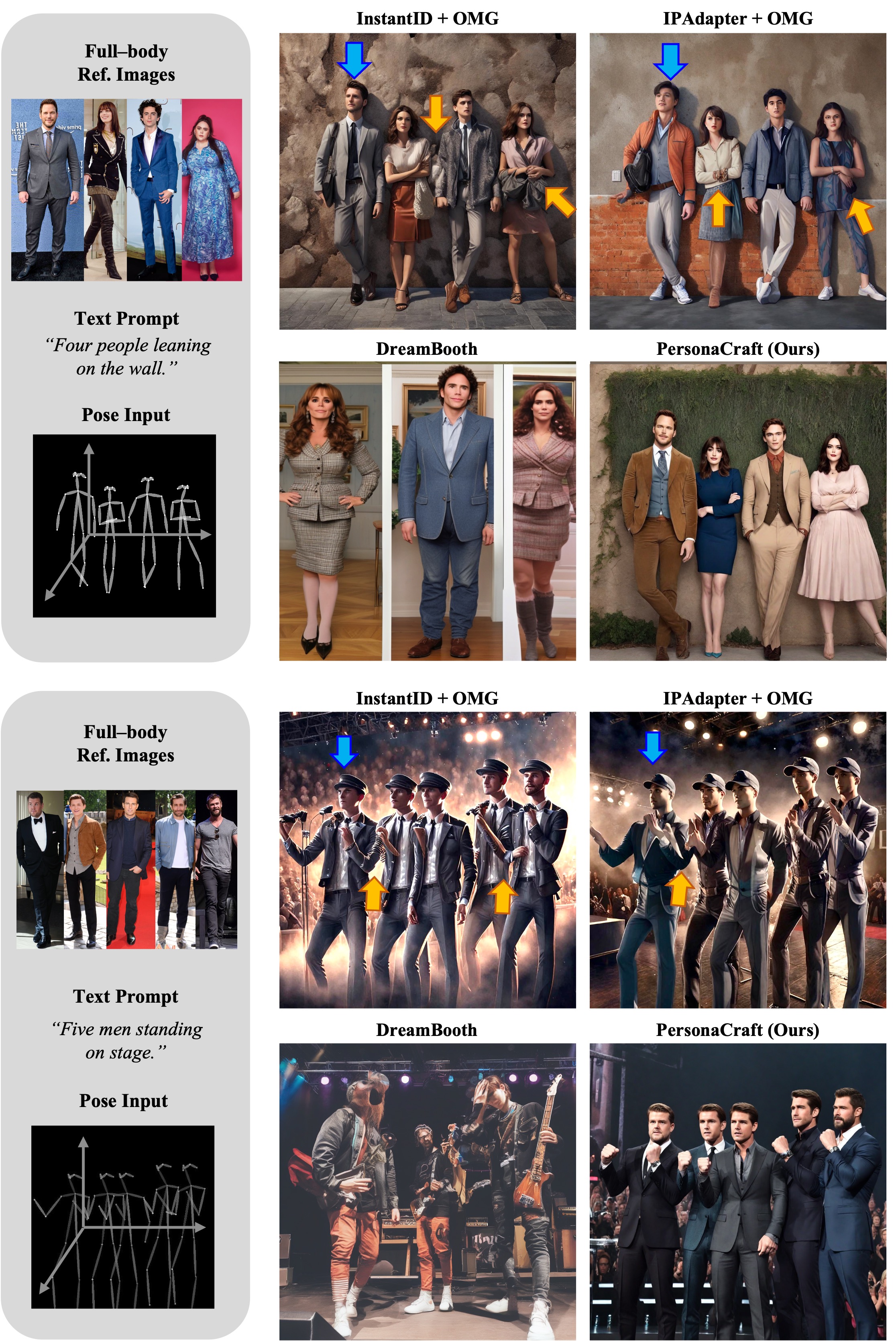}
    \caption{Additional comparison of generated images. Yellow arrows highlight anatomical issues in InstantID+OMG~\cite{wang2024instantid, kong2025omg} and IPAdapter+OMG~\cite{ye2023ip-adapter, kong2025omg} due to 2D pose limitations. DreamBooth~\cite{ruiz2023dreambooth} shows severe distortions and clothing mismatches. PersonaCraft excels in face identity, body shape consistency, and naturalness while being over twice as fast as OMG-based methods~\cite{wang2024instantid, ye2023ip-adapter, kong2025omg}.}
    \label{fig_supp_more_results2}
\end{figure*}

\noindent \textbf{Additional Qualitative Comparison.}
As shown in Fig.~\ref{fig_supp_more_results1} and ~\ref{fig_supp_more_results2}, our proposed method demonstrates significant advantages over existing approaches. Notably, methods like InstantID+OMG~\cite{wang2024instantid, kong2025omg} and IPAdapter+OMG~\cite{ye2023ip-adapter, kong2025omg}, which rely on 2D skeleton-based pose conditioning, exhibit severe anatomical inaccuracies in challenging scenarios (highlighted by yellow arrows). These issues stem from the inherent limitations of 2D pose representations, which struggle to handle overlapping body parts and intricate interactions effectively. Blue arrows highlight cases for verifying correct body shape preservation, where our approach maintains accurate body structure and pose fidelity while delivering superior performance in both face identity preservation and body shape consistency.

DreamBooth~\cite{ruiz2023dreambooth}, on the other hand, suffers even more pronounced issues due to its lack of pose guidance. This leads to severe anatomical distortions and, in some cases, the complete omission of individuals in multi-person scenes. Additionally, DreamBooth struggles with clothing-body shape displacement, where clothing styles are directly transferred without adapting to the individual’s body shape.

These findings further underscore the robustness and versatility of PersonaCraft, making it a state-of-the-art solution for personalized image generation in complex, real-world scenarios.

\begin{figure*}[!t]
    \centering
    \includegraphics[width=.7\textwidth]{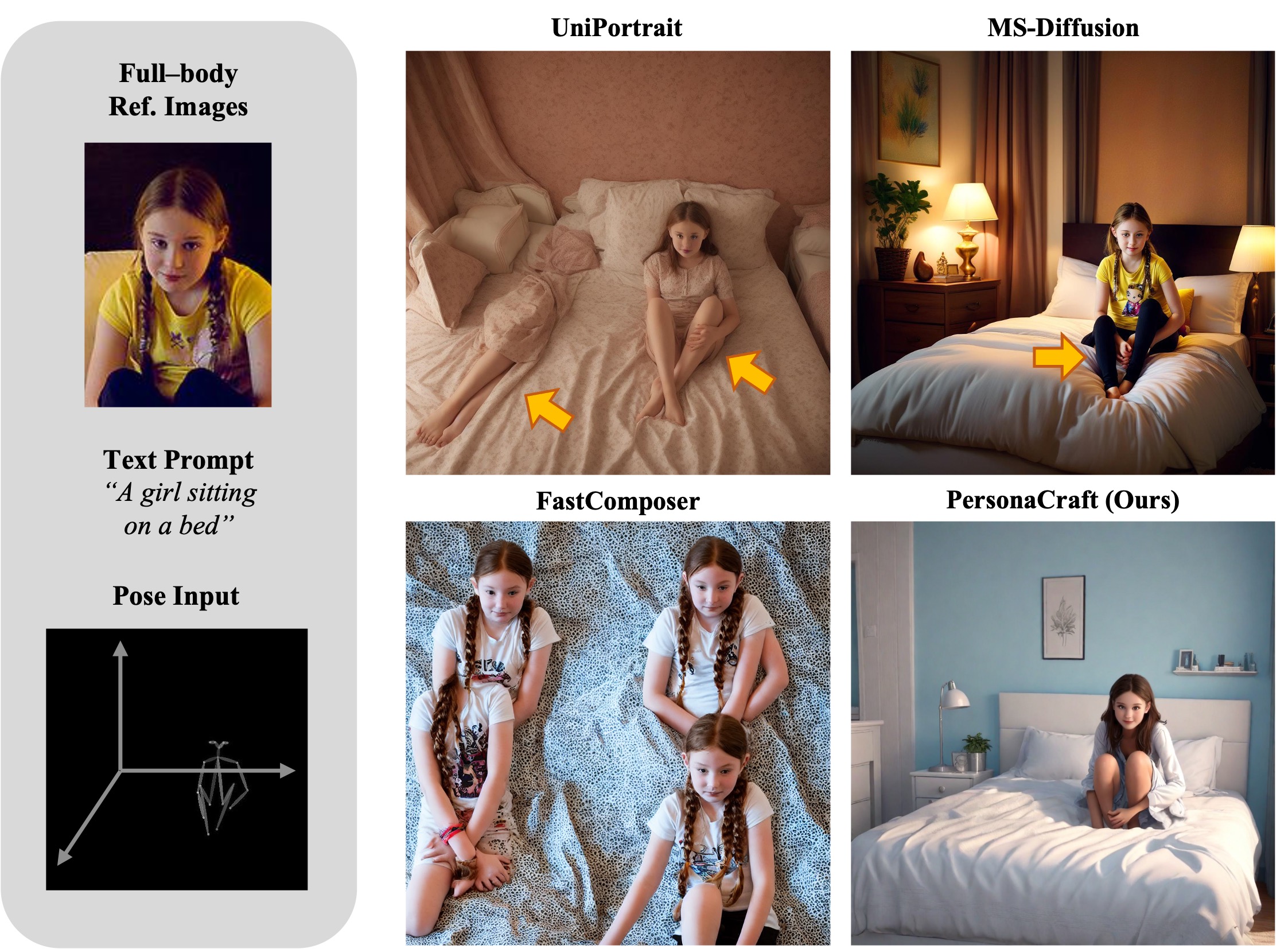}
    \caption{Comparison of PersonaCraft with UniPortrait~\cite{he2024uniportrait}, MS-Diffusion~\cite{wang2024ms}, and FastComposer~\cite{xiao2024fastcomposer}. Yellow arrows show anatomical inconsistencies in poses and occlusions. MS-Diffusion copies clothing without proper adjustment.PersonaCraft excels in face identity, body shape consistency, and naturalness compared to the baselines.}
    \label{fig_supp_additional_baselines1}
\end{figure*}

\noindent \textbf{Comparison with Additional Baselines.}
We compared PersonaCraft against other baselines, including, UniPortrait~\cite{he2024uniportrait}, MS-Diffusion~\cite{wang2024ms}, and FastComposer~\cite{xiao2024fastcomposer}. While these methods share similar capabilities, they are not fully suited for our benchmark, making direct comparisons challenging. As shown in (Fig.\ref{fig_supp_additional_baselines1}), yellow arrows highlight anatomical inconsistencies in complex poses and occluded scenarios due to reliance on 2D pose representations or the absence of pose control. PersonaCraft, in contrast, generates anatomically accurate and natural images under these
conditions. 

Additionally, MS-Diffusion copies clothing directly from
full-body references without proper displacement. PersonaCraft integrates personalized body shapes and clothing
displacement, maintaining consistency and realism.

These results highlight PersonaCraft’s superiority in generating accurate, identity-consistent images and handling
occlusions and diverse poses with exceptional naturalness
and customization.

\subsection{Pose-Controlled Multi-Human Scene Generation}
\begin{figure*}[!h]
    \centering
    \includegraphics[width=.7\textwidth]{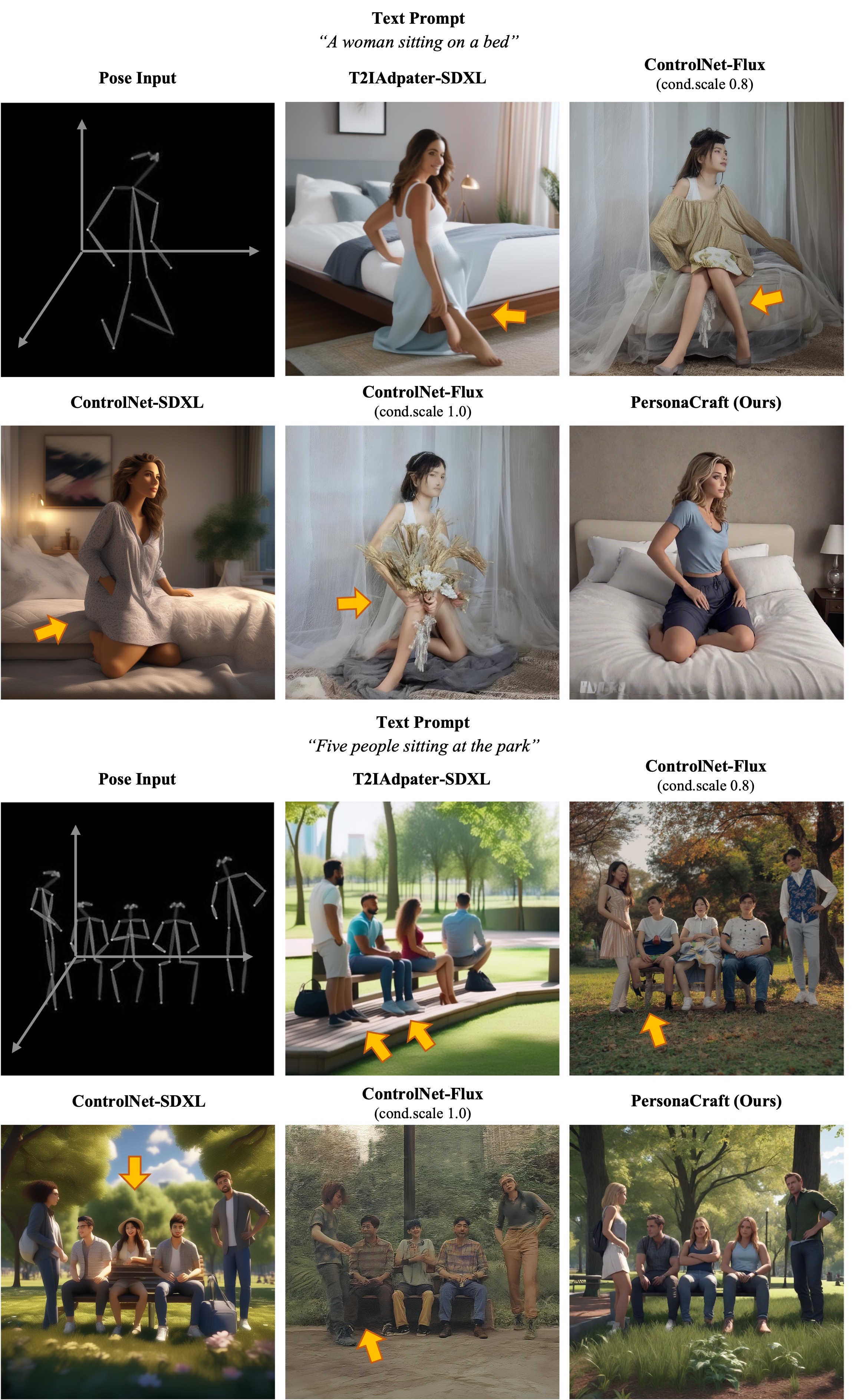}
    \caption{Qualitative comparison of 3D-aware pose conditioning for multi-human generation, covering both single and multi-human scenarios. PersonaCraft achieves superior alignment with the input pose while effectively handling occlusions, allowing for natural human anatomy to be maintained even in complex multi-human interactions. Our method outperforms baselines in preserving identity, body shape, and overall human realism. \textcolor{yellow}{yellow} arrows highlight unnatural anatomical structures.}
    \label{fig_supp_more_pose_results1}
\end{figure*}

\begin{figure*}[!h]
    \centering
    \includegraphics[width=.7\textwidth]{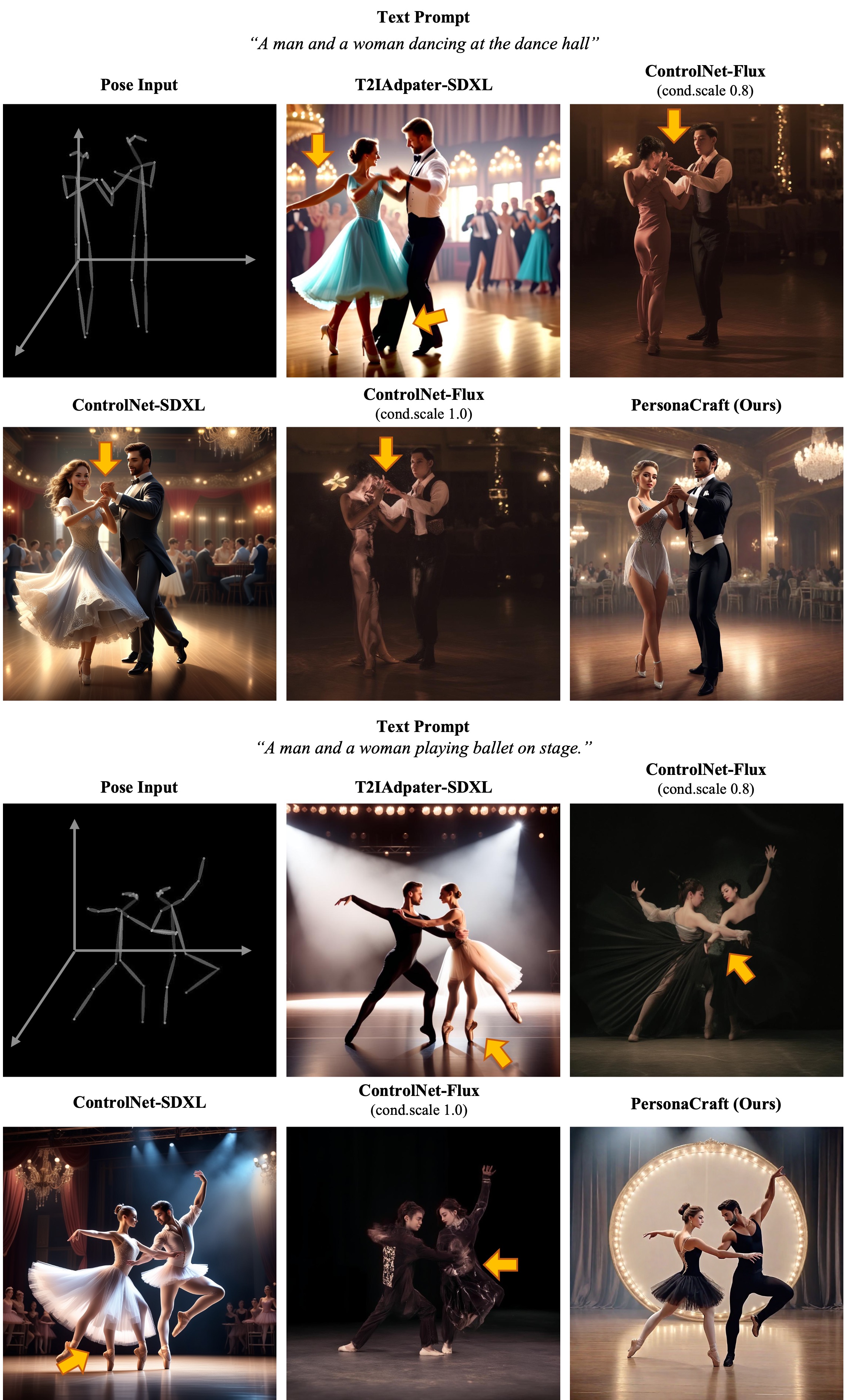}
    \caption{Qualitative comparison of 3D-aware pose conditioning for multi-human generation, covering both single and multi-human scenarios. PersonaCraft achieves superior alignment with the input pose while effectively handling occlusions, allowing for natural human anatomy to be maintained even in complex multi-human interactions. Our method outperforms baselines in preserving identity, body shape, and overall human realism. \textcolor{yellow}{yellow} arrows highlight unnatural anatomical structures.}
    \label{fig_supp_more_pose_results2}
\end{figure*}

\begin{figure*}[!t]
    \centering
    \includegraphics[width=.7\textwidth]{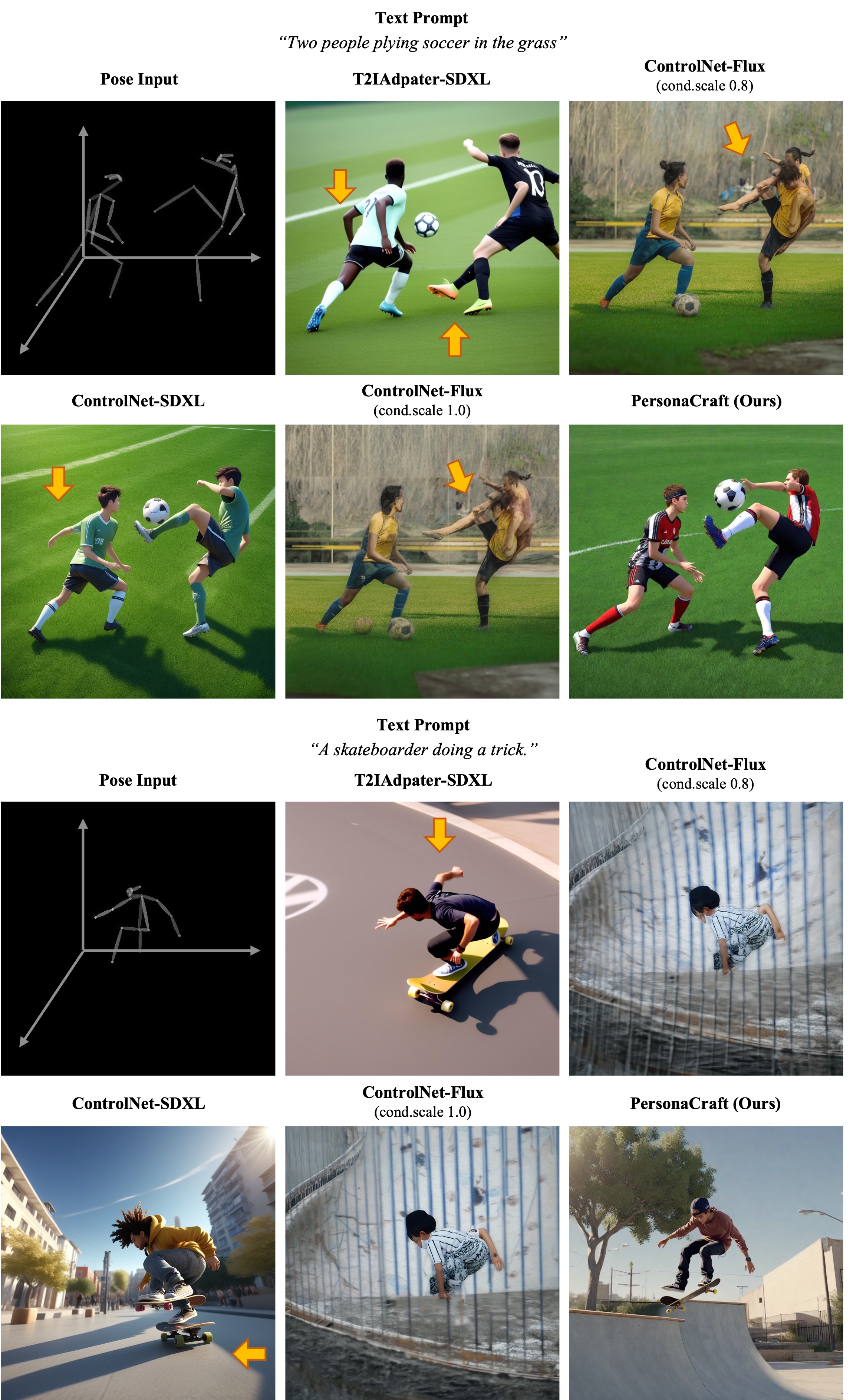}
    \caption{Qualitative comparison of 3D-aware pose conditioning for multi-human generation, covering both single and multi-human scenarios. PersonaCraft achieves superior alignment with the input pose while effectively handling occlusions, allowing for natural human anatomy to be maintained even in complex multi-human interactions. Our method outperforms baselines in preserving identity, body shape, and overall human realism. \textcolor{yellow}{yellow} arrows highlight unnatural anatomical structures.}
    \label{fig_supp_more_pose_results3}
\end{figure*}

We assess pose-controlled multi-human scene generation by comparing PersonaCraft with existing methods in both single-human and multi-human contexts. As shown in Fig.~\ref{fig_supp_more_pose_results1},~\ref{fig_supp_more_pose_results2}, and~\ref{fig_supp_more_pose_results3}, our proposed method demonstrates significant advantages over existing approaches. Notably, methods that rely on 2D pose conditioning (e.g., T2IAdapter-SDXL\cite{mou2023t2i}, ControlNet-SDXL~\cite{zhang2023adding,podell2023sdxl}) struggle with occlusion handling, resulting in misaligned poses and distorted structures. ControlNet-Flux~\cite{flux2024,zhang2023adding} shows instability: at a lower conditioning scale (0.8), it fails to preserve anatomical coherence, while at a higher scale (1.0), pose accuracy improves but image fidelity decreases. 
In contrast, PersonaCraft successfully balances pose accuracy, body realism, and image fidelity, highlighting the effectiveness of our 3D-aware pose conditioning.

\subsection{PersonaCraft with Stylization}
\begin{figure*}[!t]
    \centering
    \includegraphics[width=1\textwidth]{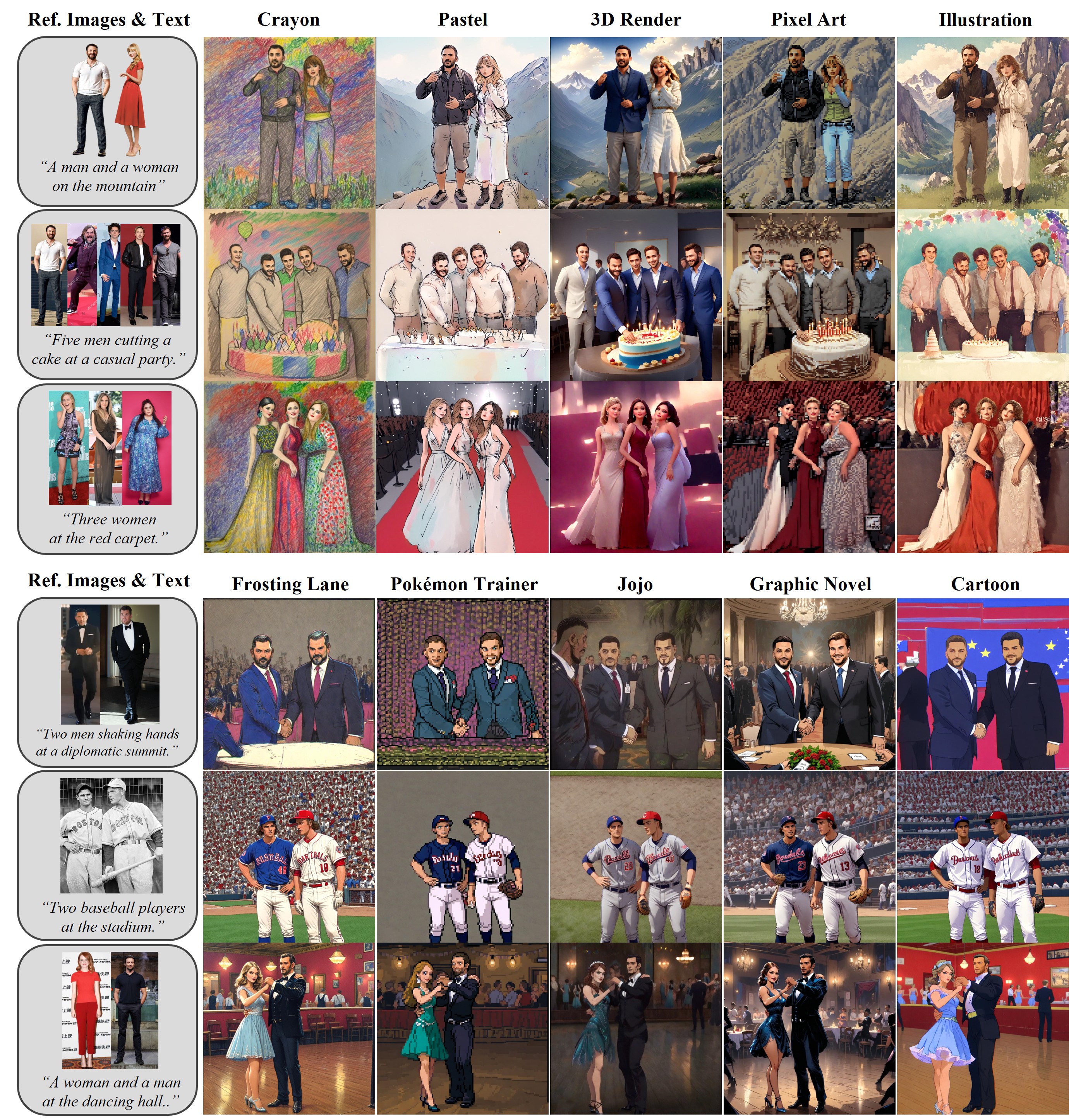}
    \vspace{-1em}
    \caption{Results of combining PersonaCraft with various style LoRAs, showcasing adaptability to styles like Pastel, JoJo, and Pokémon Trainer. Some styles alter facial and body identities due to their bias, while producing visually impressive outcomes.}
    \label{fig_supp_style}
\end{figure*}

The proposed method is a plug-and-play approach, making it compatible with various style-specific LoRAs. To evaluate its effectiveness, we conducted experiments combining PersonaCraft with diverse style LoRAs, including Crayon~\cite{crayonstylelora}, Pastel~\cite{softpastelanime}, 3D Render~\cite{3drenderstylexl}, Pixel Art~\cite{pixelportraits192}, Illustration~\cite{paintinglight}, Frosting Lane~\cite{frostinglane}, Pokémon Trainer~\cite{pokemontrainer}, JoJo~\cite{jojostylelora}, Graphic Novel~\cite{graphicnovelillustration}, and Cartoon~\cite{geminianime}. The results, shown in Fig.~\ref{fig_supp_style}, highlight the method's ability to adapt to different styles effectively. Notably, styles such as Pastel, Illustration, JoJo, and Pokémon Trainer introduce changes in facial and body characteristics, occasionally altering perceived identity,  due to their bias. Nevertheless, the outcomes remain visually compelling and demonstrate the versatility of our approach.

\subsection{Versatility of SCNet}
\begin{figure*}[!t]
    \centering
    \includegraphics[width=0.8\textwidth]{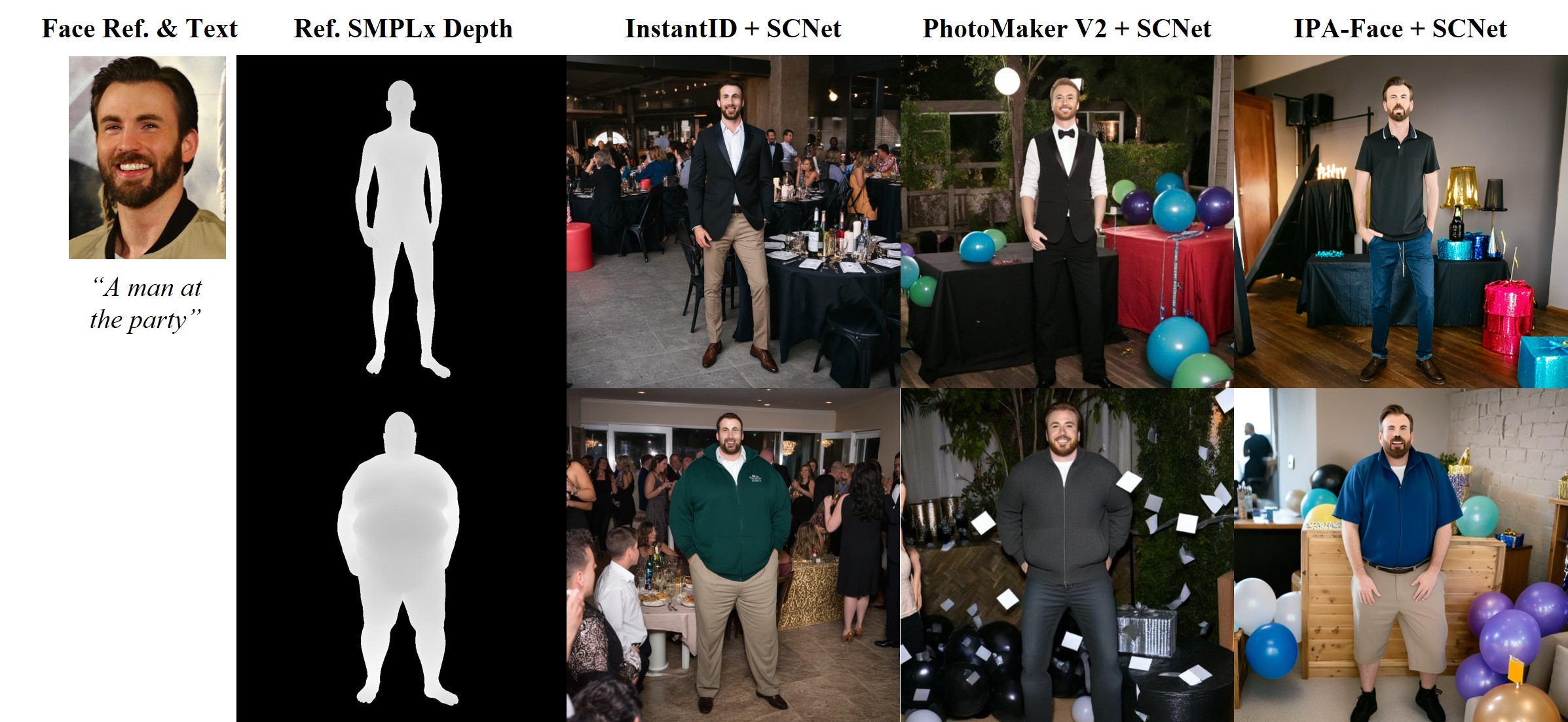}
    \vspace{-.5em}
    \caption{Integration of SCNet with face personalization models like InstantID~\cite{wang2024instantid}, PhotoMaker V2~\cite{li2023photomaker} and IPAdapter-Face~\cite{ye2023ip-adapter} achieves robust full-body customization, with slight variations by face module.}
    \label{fig_supp_versatile_scnet}
\end{figure*}

To demonstrate the versatility of SCNet, we present results combining SCNet with various face identity personalization models, including InstantID~\cite{wang2024instantid}, PhotoMaker V2~\cite{li2023photomaker}, and IPAdapter-Face~\cite{ye2023ip-adapter}. As shown in Fig.~\ref{fig_supp_versatile_scnet}, SCNet enables robust body shape personalization and pose control when paired with these face models, achieving comprehensive full-body personalization and user-defined body shape adjustments. Notably, face personalization varies slightly depending on the chosen face module.

\subsection{Ablation Study on Conditioning Scale}
\begin{figure*}[!t]
    \centering
    \includegraphics[width=0.8\textwidth]{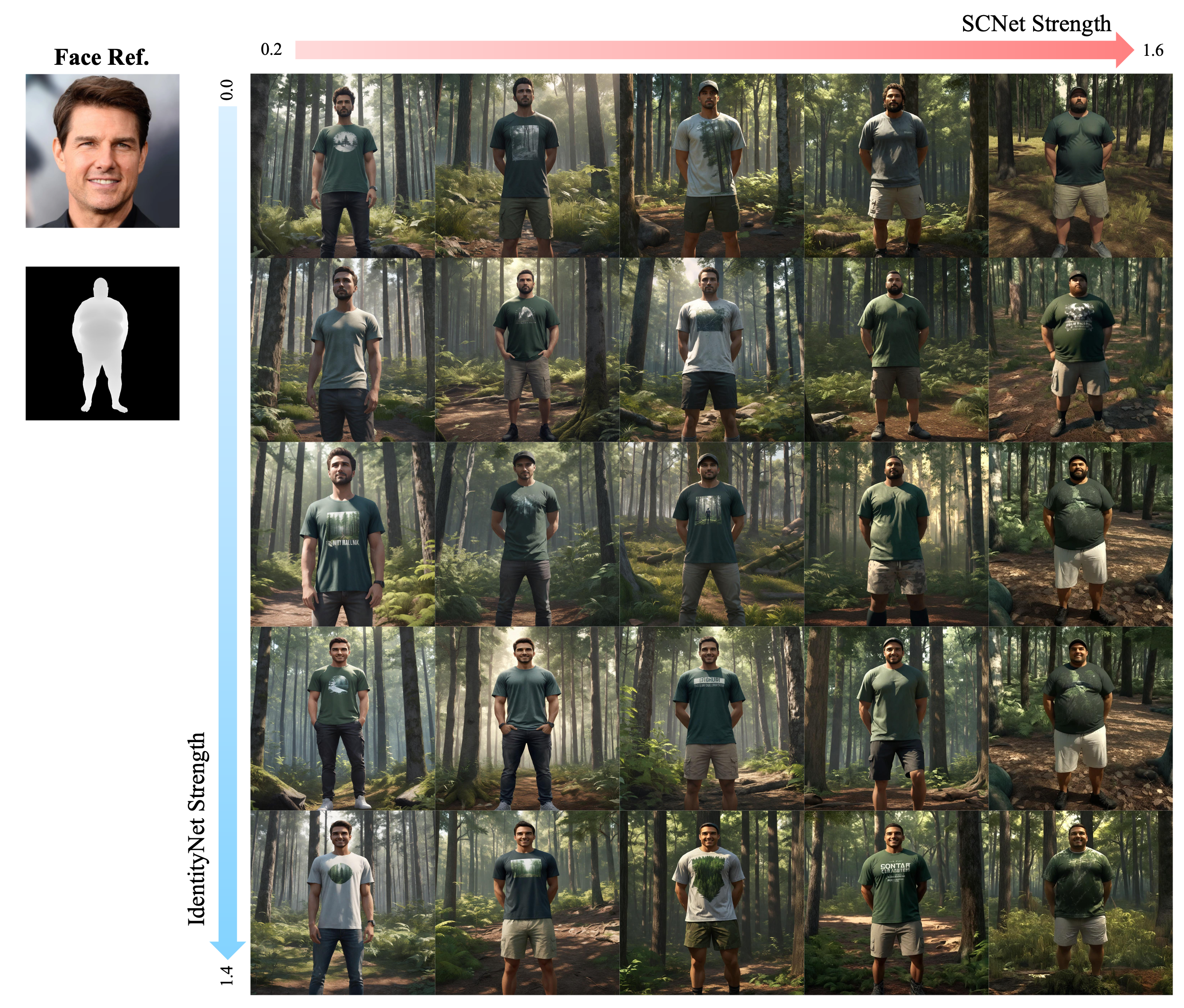}
    \caption{Ablation study on the conditioning scales of IdentityNet and SCNet, demonstrating improved identity preservation for face and body shape as the conditioning scales increase.}
    \label{fig_supp_ablation}
\end{figure*}

We analyze the effect of the conditioning scales of IdentityNet and SCNet on identity preservation when provided with face references and reference body shapes (SMPLx depth). As shown in Fig.~\ref{fig_supp_ablation}, when the conditioning scale is set to 0 for both modules, the generated face and body shapes differ significantly from the reference. This indicates insufficient guidance from the reference inputs.

As the conditioning scales for IdentityNet and SCNet increase, the generated images progressively resemble the reference face and body shape. This improvement demonstrates the critical role of conditioning strength in aligning the generated outputs with the given references. Optimal conditioning scales enable PersonaCraft to faithfully preserve both facial and body shape identities, ensuring high-quality personalization and consistency.

\subsection{Ablation Study on Body Shape Parameters}
For full-body personalized image generation, we extract the body shape parameters of the character to be personalized and use them for SMPLx rendering, which serves as the conditions for SCNet. In Tab.~\ref{tab:ablation_body}, we analyze the impact of incorporating the body shape parameters in this process. 
Using body shape parameters enhances body shape preservation during personalization. This indicates that leveraging the body shape parameters enables the generation of personalized images that more accurately reflect the character’s true physique.

\begin{table}[!ht]\centering
\caption{Evaluation of body shape preservation with and without the use of the body shape parameter.}\label{tab: }
\scriptsize
\vspace{-1em}
\label{tab:ablation_body}
\centering
\begin{adjustbox}{width=0.65\linewidth}
\begin{tabular}{ccccc}\toprule
&Single &Multi &Total \\\midrule
w/o body shape &0.615 &0.520 &0.539 \\
w/ body shape &\textbf{0.630} &\textbf{0.548} &\textbf{0.615} \\
\bottomrule
\end{tabular}
\end{adjustbox}
\vspace{-1em}
\end{table}
 

\subsection{Ablation Study on Occlusion-aware 3D pose \& Shape conditioning}
\begin{figure*}[!t]
    \centering
    \includegraphics[width=0.9\textwidth]{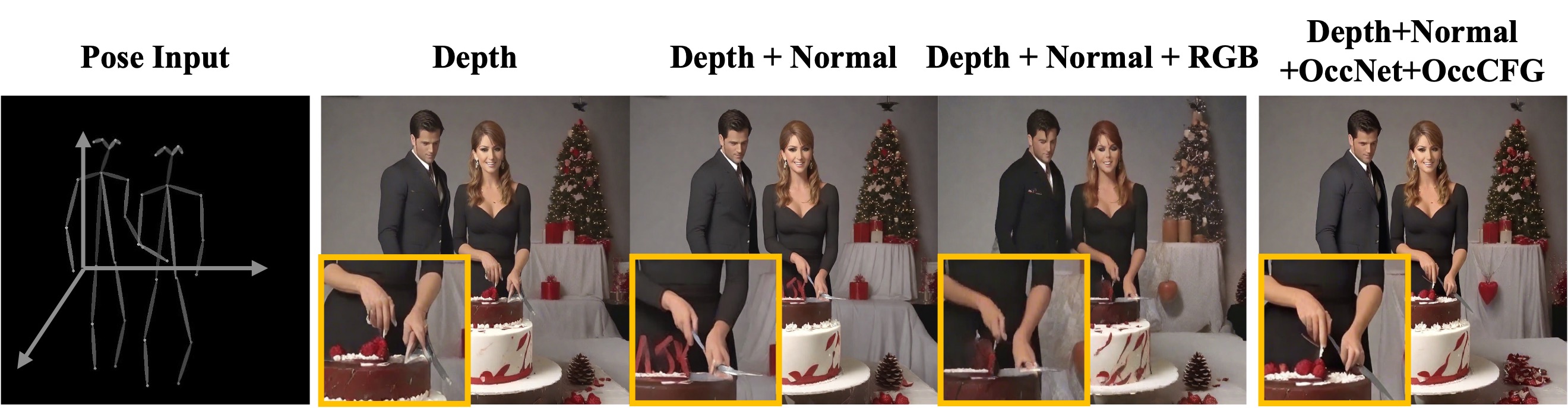}
    \caption{Ablation study on occlusion-aware 3D pose \& shape conditioning. The combination of depth and normal as conditioning inputs for SCNet achieves the best generation performance in occluded or complex regions. While using SCNet faces issues preserving pose structure in fine-grained occluded regions, adding OccNet and OccCFG effectively addresses these problems.}
    \label{fig_supp_pose_analysis}
\end{figure*}

\noindent \textbf{Comparison of 3D Pose Representations for SCNet.}
In Fig.~\ref{fig_supp_pose_analysis}, we compare different combinations of SMPLx rendering-depth, normal, and RGB rendering-as conditioning inputs for SCNet. Using both depth and normal enables the model to leverage occlusion cues from depth and surface orientation information from normal. This leads to improved generation performance in occluded or complex body regions compared to using depth alone. However, incorporating RGB rendering in addition to depth and normal degrades image quality due to the use of multiple ControlNets for the same region. Therefore, we adopt the combination of depth and normal as base conditioning for SCNet.

\noindent \textbf{Effectiveness of OccNet and OccCFG.}
Also, the effect of our occlusion-aware 3D pose \& shape conditioning components is analyzed in Fig.~\ref{fig_supp_pose_analysis}. Using only 2D pose leads to structural inconsistencies, while relying solely on 3D pose (SCNet) struggles with fine-grained occluded regions. Our full model (SCNet + OccNet + OccCFG) effectively preserves pose structure while handling occlusions.

\subsection{Efficiency Analysis}

As analyzed in Tab.~\ref{tab:inference_time}, we compare the inference times for multi-identity personalized synthesis across different methods, specifically for generating images with three distinct identities. Textual Inversion~\cite{gal2022image} and DreamBooth~\cite{ruiz2023dreambooth}, which rely on optimization-based personalization, require a batch size of 4 and 500 optimization steps per identity. This process results in significantly longer inference times, making these methods highly inefficient for real-time applications.

On the other hand, methods based on OMG~\cite{kong2025omg}, which utilize a two-stage process to generate images, also require over twice the amount of time compared to our approach. In contrast, PersonaCraft performs inference efficiently, generating personalized images with substantially lower computation time while maintaining high-quality results. This stark difference highlights the efficiency advantage of our method, especially when generating multiple identities in a single synthesis process.

\begin{table}[!ht]\centering
\vspace{-.5em}
\caption{Inference times for multi-identity personalized synthesis.}\label{tab:inference_time}
\scriptsize
\vspace{-1em}
\label{tab:user}
\centering
\begin{adjustbox}{width=0.6\linewidth}
\begin{tabular}{ccc}\toprule
{Method} & {Total Time (secs)} \\\midrule
Text Inversion &1636.15 \\
DreamBooth &770.713 \\
InstantID + OMG &46.94 \\
IPAdapter + OMG &44.62 \\
IPA-Face + OMG &35.46 \\
\textbf{PersonaCraft (ours)} & \textbf{17.25} \\
\bottomrule
\end{tabular}
\end{adjustbox}
\vspace{-1.5em}
\end{table}



\subsection{Additional Results with Baselines Finetuned on our Training Dataset}
We further compare PersonaCraft with key baselines finetuned on our training dataset: InstantID + OMG, IPAdapter + OMG, and IPA-Face + OMG for personalized multi-human scene generation, and ControlNet-SDXL for pose-controlled multi-human scene generation.

\noindent \textbf{Personalized Multi-Human Scene Generation.}
\begin{table*}[!t]\centering
\caption{
Additional comparison of baseline models fine-tuned on our training dataset (MPII). Quantitative evaluation of multi-human personalization across face identity, body shape preservation, pose accuracy, text alignment, and image quality. ‘Single’ denotes personalization for a single individual, while ‘Multi’ refers to cases with multiple identities (2–5), with ‘Total’ representing the averaged results. (*: fine-tuned on our training dataset.)
}
\label{tab:personalization_supp}
\scriptsize
\vspace{-1em}
\centering
\begin{adjustbox}{width=0.9\linewidth}
\begin{tabular}{cccccccccccc}
\toprule
\multirow{2}{*}[-0.5em]{\shortstack[c]{Multi-Human\\Personalization}} 
& \multicolumn{3}{c}{Face ID preservation\textuparrow} 
& \multicolumn{3}{c}{Body shape preservation\textuparrow} 
& \multicolumn{2}{c}{Pose} 
& \multicolumn{1}{c}{Text} 
& \multicolumn{2}{c}{Image quality}\\

\cmidrule(lr){2-4} \cmidrule(lr){5-7} \cmidrule(lr){8-9} \cmidrule(lr){10-10} \cmidrule(lr){11-12}
& Single & Multi & Total & Single & Multi & Total 
& MPJPE (3D)\textdownarrow & AP-0.5 (2D)\textuparrow & CLIP sim \textuparrow & IS \textuparrow & KID \textdownarrow \\
\midrule
InstantID + OMG$^*$ & 0.418 & 0.220 & 0.252 & 0.563 & 0.427 & 0.448 & 85.624 & 0.333 & 0.264 & 3.809 & 0.0996 \\
IPAdapter+ OMG$^*$  & 0.204 & 0.155 & 0.162 & 0.606 & 0.45 & 0.472 & 84.893 & 0.362 & 0.267 & 3.736 & 0.0984 \\
IPA-Face + OMG$^*$   & 0.350 & 0.181 & 0.207 & 0.568 & 0.429 & 0.451 & 86.373 & 0.355 & 0.265 & 3.930 & 0.0974 \\
\textbf{Ours} & \textbf{0.421} & \textbf{0.298} & \textbf{0.317} & \textbf{0.630} & \textbf{0.548} & \textbf{0.560} & \textbf{60.654} & \textbf{0.506} & \textbf{0.273} & \textbf{4.238} & \textbf{0.0931} \\
\bottomrule
\end{tabular}
\end{adjustbox}
\vspace{-2em}
\end{table*}
Although the baselines are fine-tuned on our training dataset, Tab.~\ref{tab:personalization_supp} shows that our method consistently outperforms baselines in identity and body shape preservation. It achieves the lowest MPJPE (3D) and highest AP (2D), indicating superior alignment with input poses. Additionally, PersonaCraft surpasses baselines in IS and KID, demonstrating enhanced perceptual quality and text-image coherence.

\noindent \textbf{Pose-Controlled Multi-Human Scene Generation.}
\begin{table}[!t]\centering
\caption{
Additional comparison of the baseline fine-tuned on our training dataset (MPII).
Quantitative evaluation of pose-controlled human generation in terms of pose accuracy, text alignment, and image quality. (*: fine-tuned on our training dataset.)}
\label{tab:pose_controlled_supp}
\scriptsize
\vspace{-1em}
\centering
\begin{adjustbox}{width=1\linewidth}
\begin{tabular}{cccccc}
\toprule
\multirow{2}{*}[-0.5em]{\shortstack[c]{Pose-Controlled\\Human Generation}} 
& \multicolumn{2}{c}{Pose} 
& \multicolumn{1}{c}{Text} 
& \multicolumn{2}{c}{Image quality}\\
\cmidrule(lr){2-3} \cmidrule(lr){4-4} \cmidrule(lr){5-6}
& MPJPE (3D) \textdownarrow & AP-0.5 (2D) \textuparrow & CLIP sim \textuparrow & IS \textuparrow & KID \textdownarrow \\
\midrule
ControlNet-SDXL$^*$ & 100.817 & 0.313 & \textbf{0.281} & 3.407 & 0.122 \\
\textbf{Ours}  & \textbf{62.647} & \textbf{0.495} & 0.274 & \textbf{4.114} & \textbf{0.091} \\
\bottomrule
\end{tabular}
\end{adjustbox}
\vspace{-1.5em}
\end{table}
Although the baselines are fine-tuned on our training dataset, Tab.~\ref{tab:pose_controlled_supp} shows that PersonaCraft achieves the lowest MPJPE (3D) and highest AP-0.5 (2D), demonstrating superior pose alignment and keypoint localization.

\subsection{Failure Cases}
\begin{figure*}[!t]
    \centering
    \includegraphics[width=0.6\textwidth]{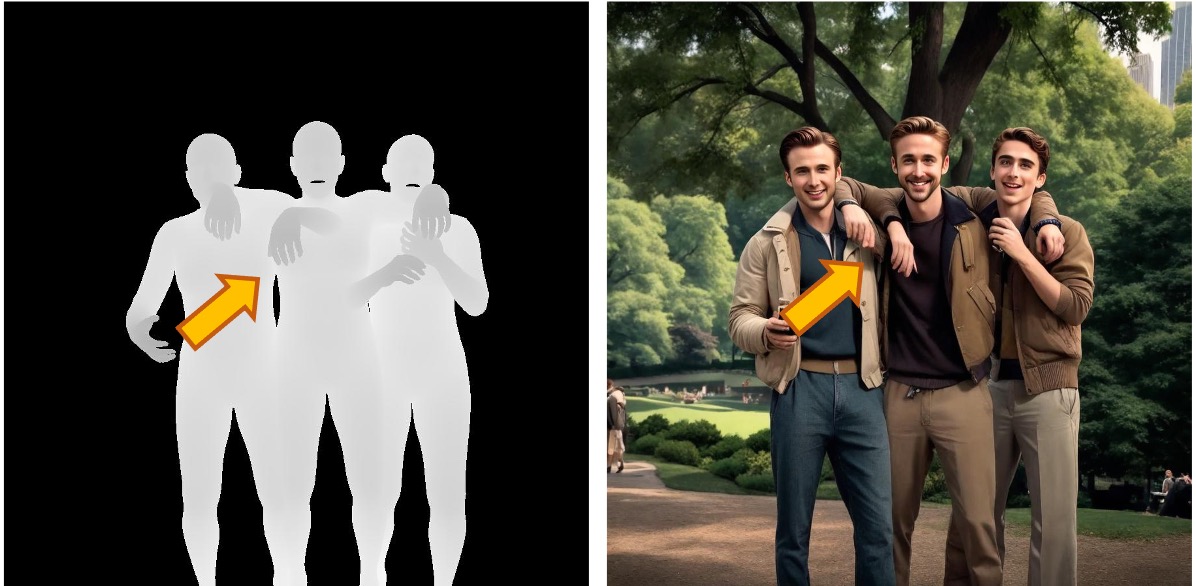}
    \caption{Failure cases. The accuracy of 3D human model fitting depends on the fitting algorithm used, and variations in fitting quality, particularly when the reference data is incomplete or inaccurate, can impact the final output.}
    \label{fig_supp_failure}
\end{figure*}

While our method is versatile and can be applied to other ControlNet models, the performance of our face personalization depends significantly on the underlying face network. Additionally, the accuracy of 3D human model fitting is dependent on the performance of the fitting algorithm used. Variations in the quality of the fitting process may impact the final output, especially in cases where the reference data is incomplete or inaccurate as presented in Fig.~\ref{fig_supp_failure}.

\noindent (Additional Experimental details are continued in Sec.~\ref{supp:details})

\clearpage
\clearpage


\section{Additional Experimental details}
\label{supp:details}
\subsection{Additional Implementation Details}

\noindent \textbf{Additional Details on Training Dataset.} 
To account for occlusion scenarios, we balance the MPII~\cite{andriluka20142d} dataset with a 2:1 ratio of single-person to multi-person images. Depth clipping is applied during depth rendering to retain only values below 5, ensuring consistent quality.
After preprocessing, we curate a final dataset of 6,348 image-text-SMPLx-parameter pairs. This carefully curated dataset enables robust model training with diverse 3D human poses, complex interactions, and detailed human parameters such as body shape and pose conditioning.

\noindent \textbf{Details on Training of SCNet and OccNet.}
We base our SCNet on \texttt{controlnet-union-sdxl-1.0}~\cite{controlnet_sdxl_2023} and fine-tune it for SMPLx~\cite{SMPL-X:2019} depth-normal-ocluded edge conditioning. The architecture supports more than 10 control types for high-resolution text-to-image generation, with depth selected as the control type in our implementation.

We utilize 3D poses represented by SMPLx parameters, which include 55 joints (22 body, 1 jaw, 2 eyes, and 30 hands) along with camera parameters (intrinsic and extrinsic). These parameters generate a vertex- and face-based mesh that we render as SMPLx depth maps.

For depth edge extraction, we employed the Canny edge detector for more robust edge extraction instead of thresholding the spatial partial gradient of depth with \(\tau\), using a low threshold of 5 and a high threshold of 15.

\noindent \textbf{Details on Full-Body Personalized Image Synthesis.}  
We adopt MultiHMR~\cite{multi-hmr2024} as our SMPLx fitting method. MultiHMR is a single-shot model that reconstructs 3D human meshes from a single RGB image, leveraging the SMPLx parametric model to predict full-body meshes, including hands and facial expressions, with 3D localization in the camera coordinate system. The body shape parameters, \(\bm{\beta}\), are represented as 10-dimensional vectors, each scaled by orthonormal shape displacement components.  

For facial identity processing, we employ the \texttt{antelopev2} facial detection and recognition models from InsightFace~\cite{insightface} to extract 512-dimensional face identity embeddings, \(\bm{f}\), from human images.  

To enhance the visual quality of human-centric scenes, we utilize the \texttt{YamerMIX-v8} variant of SDXL.  
For face identity personalization, we incorporate IdentityNet from InstantID~\cite{wang2024instantid}, which enables instant, zero-shot, identity-preserving image generation. IdentityNet enforces strong semantic and weak spatial conditions by integrating facial and landmark images with textual prompts to guide the generation process.  

Following InstantID, we use five key facial landmarks (two for the eyes, one for the nose, and two for the mouth) as spatial control signals, providing a more generalized constraint than detailed key points.

\noindent \textbf{Details on Dual-Pathway Body Shape Personalization.}  
In this method, a CLIP~\cite{radford2021learning} -based classifier is employed to extract body shape attributes in the form of text descriptions. These descriptions categorize the body type into various categories, such as "overweight," "muscular," "fat," etc. This is achieved by using a combination of CLIP, which bridges the gap between vision and text, and specific regional prompting techniques.

The body shape information is then used in Regional Diffusion, a concept derived from MultiDiffusion~\cite{bar2023multidiffusion}, where each diffusion timestep involves conditioning on both the body pose and a corresponding text description about the individual’s body shape. The process operates on each person’s instance throughout multiple diffusion timesteps, ensuring that the shape-specific features are incorporated and aligned with the pose dynamics.

Incorporating these shape attributes into the diffusion process allows the model to better represent personalized body shapes in the generation process, resulting in a more accurate and detailed synthesis of human shapes. This integration is achieved through the use of regional prompting, where at each timestep, the model is conditioned on both the body pose and the specific body shape description to refine the body shape in the generated instance. This process is further integrated with SCNet, a network that helps guide the shape refinement and personalization.

\noindent \textbf{User-Defined Body Shape Control.}  
PersonaCraft enables user-defined body shape control, allowing adjustments based on user preferences:  
\textit{1) Reference-based control}: The target body shape parameter, $\bm{\beta}_{\text{target}}$, is obtained from a reference image via SMPLx fitting.  
\textit{2) Interpolation/extrapolation-based control}: Given two reference body shape parameters, $\bm{\beta}_1$ and $\bm{\beta}_2$, the target shape is computed as 
$\bm{\beta}_{\text{target}} = \gamma \bm{\beta}_1 + (1 - \gamma) \bm{\beta}_2$,
where $\gamma$ controls the interpolation/extrapolation ratio. The resulting $\bm{\beta}_{\text{target}}$ .

\subsection{Details on Metrics}
\noindent \textbf{Face identity preservation} was measured for $1\sim5$ identities following FastComposer~\cite{xiao2023fastcomposer}, using FaceNet~\cite{schroff2015facenet} for identity similarity within the face mask. The identity similarity score is computed by averaging the non-negative cosine similarity over both the number of humans and the total number of images:

\small
\begin{equation}
\mathcal{S}_{\text{face}} = \frac{1}{N_{\text{image}}} \sum_{i=1}^{N_{\text{image}}} \frac{1}{N_{\text{human}, i}} \sum_{j=1}^{N_{\text{human}, i}} \max\left(0, \frac{\bm{f}_{\text{ref}, i}^{(j)} \cdot \bm{f}_{\text{gen}, i}^{(j)}}{\|\bm{f}_{\text{ref}, i}^{(j)}\| \|\bm{f}_{\text{gen}, i}^{(j)}\|}\right)
\end{equation}
\normalsize
where \( \bm{f}_{\text{ref}, i}^{(j)} \) and \( \bm{f}_{\text{gen}, i}^{(j)} \) are the face embeddings for the \( j \)-th reference and generated identity in the \( i \)-th image, respectively. \( N_{\text{image}} \) is the total number of images, and \( N_{\text{human}, i} \) is the number of humans in the \( i \)-th image.

\noindent \textbf{Body shape preservation} was evaluated using cosine similarity between the SMPLx body shape parameters \( \bm{\beta} \) from the reference and generated instances. The score is averaged over both the number of humans and the total number of test images:
\small
\begin{equation}
\mathcal{S}_{\text{body}} = \frac{1}{N_{\text{image}}}  \sum_{i=1}^{N_{\text{image}}}  \frac{1}{N_{\text{human}, i}} \sum_{j=1}^{N_{\text{human}, i}} \frac{\bm{\beta}_{\text{ref}, i}^{(j)} \cdot \bm{\beta}_{\text{gen}, i}^{(j)}}{\|\bm{\beta}_{\text{ref}, i}^{(j)}\| \|\bm{\beta}_{\text{gen}, i}^{(j)}\|}
\end{equation}
\normalsize
where \( \bm{\beta}_{\text{ref}, i}^{(j)} \) and \( \bm{\beta}_{\text{gen}, i}^{(j)} \) are the body shape parameters for the \( j \)-th reference and generated instance in the \( i \)-th image, respectively. \( N_{\text{image}} \) is the total number of images, and \( N_{\text{human}, i} \) is the number of humans in the \( i \)-th image.

\noindent \textbf{CLIP similarity} was measured using the CLIP-L/14 model for image-text alignment. Cosine similarity was used to evaluate the alignment between the generated image and the textual description. The CLIP encoders \( \mathcal{E}_{\text{image}} \) and \( \mathcal{E}_{\text{text}} \) were used for the image and text embeddings, respectively. The alignment score is averaged over all test images:

\begin{equation}
\mathcal{S}_{\text{CLIP}} = \frac{1}{N_{\text{image}}} \sum_{i=1}^{N_{\text{image}}} \frac{\mathcal{E}_{\text{image}}(\bm{I}_{\text{gen},i}) \cdot \mathcal{E}_{\text{text}}(y_{\text{ref},i})}{\|\mathcal{E}_{\text{image}}(\bm{I}_{\text{gen},i})\| \|\mathcal{E}_{\text{text}}(y_{\text{ref},i})\|}
\end{equation}
where \( \mathcal{E}_{\text{image}}(\bm{I}_{\text{gen},i}) \) is the generated image embedding for the \( i \)-th image, and \( \mathcal{E}_{\text{text}}(y_{\text{ref},i}) \) is the reference text embedding.

\begin{figure*}[!t]
    \centering
    \includegraphics[width=0.65\textwidth]{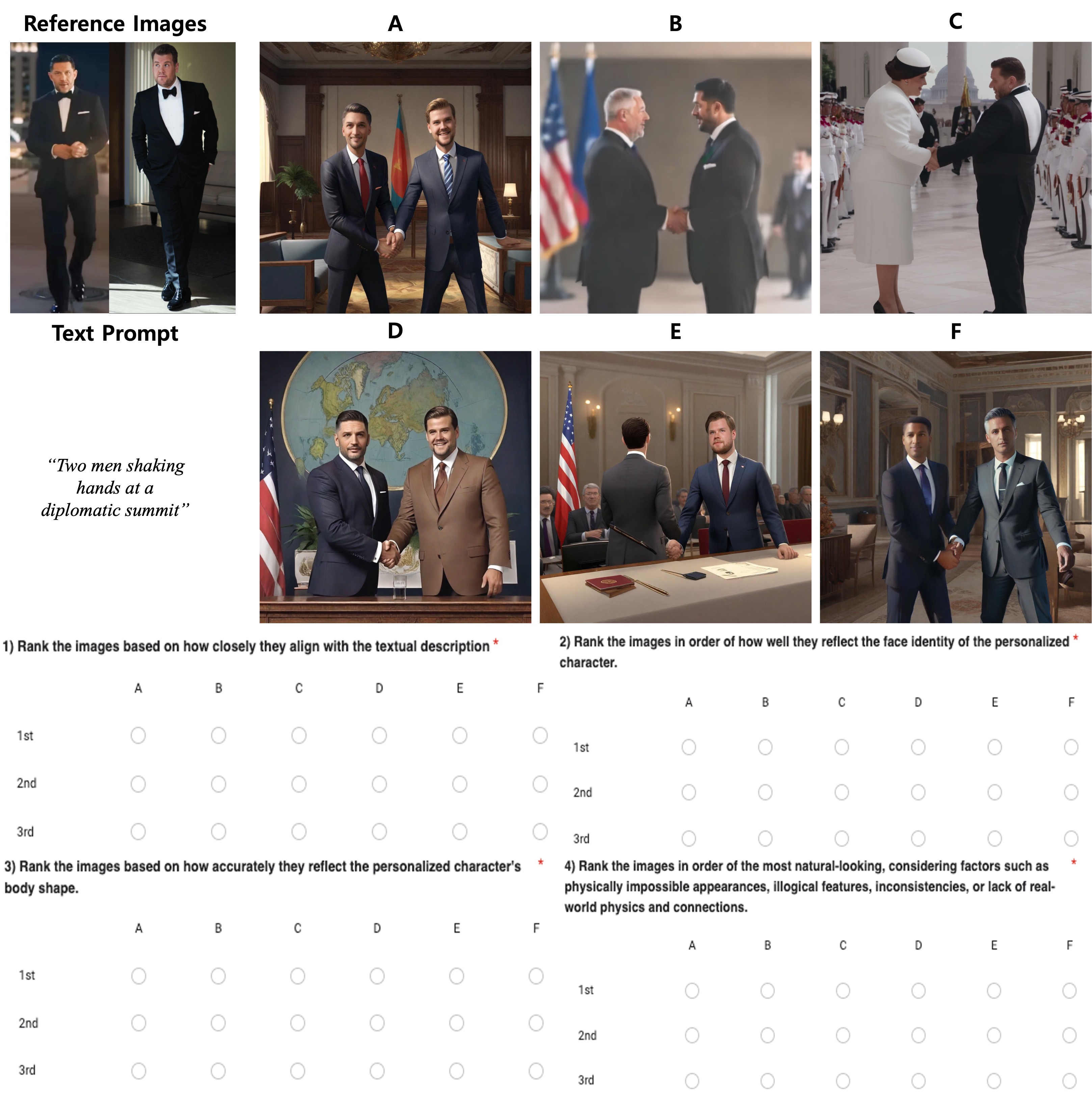}
    \vspace{-.5em}
    \caption{Example images from the user study for personalized multi-human scene generation.}
    \label{fig_supp_user_study_personalization}
\end{figure*}
\begin{figure*}[!t]
    \centering
    \includegraphics[width=0.65\textwidth]{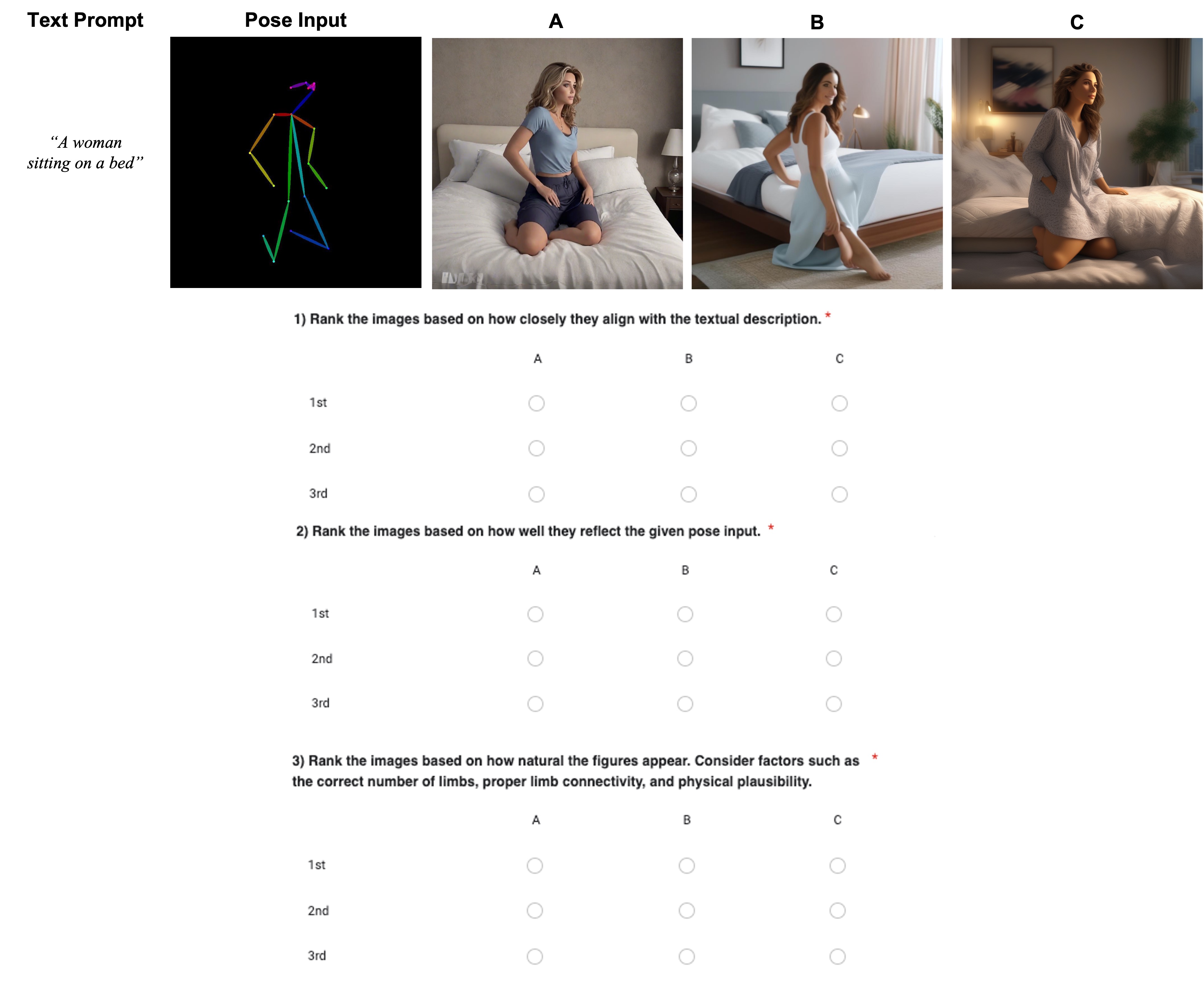}
    \vspace{-.5em}
    \caption{Example images from the user study for pose-controlled multi-human scene generation.}
    \label{fig_supp_user_study_pose}
\end{figure*}

\subsection{Details on Baselines}
To evaluate PersonaCraft, we compared it with several baselines for single-shot, multi-identity, and pose-controllable image synthesis, all implemented using SDXL~\cite{podell2023sdxl}. Key baselines include OMG~\cite{kong2025omg} for multi-concept personalization, InstantID/IPAdapter~\cite{wang2024instantid, ye2023ip-adapter} for single-shot personalization, 2D pose ControlNet~\cite{zhang2023adding}, and optimization-based methods like DreamBooth~\cite{ruiz2023dreambooth} and Texture Inversion~\cite{gal2022image}.

\noindent \textbf{OMG + InstantID/IPAdapter/IPA-Face.}  
OMG~\cite{kong2025omg} introduces a two-stage sampling solution for multi-concept personalization. The first stage handles layout generation and occlusion management, while the second stage integrates concepts using visual comprehension and noise blending. OMG can also be combined with single-concept models like InstantID without additional tuning.
For OMG+InstantID, we follow the official inference code from the InstantID repository~\cite{OMG_inference}. For OMG+InstantID and OMG+IPAdapter/IPA-Face, we replace InstantID with IPAdapter and IPA-Face, respectively, to adapt the framework for different face identity modules.

\noindent \textbf{Textual Inversion.}  
In original Textual Inversion~\cite{gal2022image}, text embeddings are optimized for user-provided visual concepts, linking them to new pseudo-words that can be seamlessly incorporated into future prompts, effectively performing an inversion into the text-embedding space.  
To enable single-reference, multi-concept personalization, we optimize a unique text embedding $ \mathcal{V}^{(i)} $ for each concept derived from a single reference image. These embeddings are paired with unique identifiers, allowing for the dynamic integration of multiple concepts into prompts during inference, facilitating multi-concept personalization.

\noindent \textbf{DreamBooth.}  
In original DreamBooth~\cite{ruiz2023dreambooth}, the model is fine-tuned with images and text prompts using a unique identifier. A prior preservation loss is applied to encourage class diversity.  
For single-reference, multi-concept personalization, we adopt DreamBooth-LoRA~\cite{ruiz2023dreambooth, hu2021lora}, where each reference image is associated with a unique $ \mathcal{M}^{(i)} $ and identifier $ \mathcal{V}^{(i)} $. These are fine-tuned based on the DreamBooth framework. During inference, both $ \mathcal{M} $ and identifiers are used simultaneously, enabling personalized, concept-specific image generation from a single reference.

\subsection{Details on User study}
\label{details_user}

\noindent \textbf{Personalized Multi-Human Scene Generation}
We conducted a user study to assess the naturalness, face identity preservation, body shape preservation, and text-image correspondence of images generated by three baseline methods (one from each group) and our method. Participants ranked the top three methods based on the following criteria:
\textit{1) Text Correspondence}: Rank the images based on how closely they align with the textual description.
\textit{2) Face Identity Preservation}: Rank the images in order of how well they reflect the face identity of the personalized character.
\textit{3) Body Shape Personalization}: Rank the images based on how accurately they reflect the personalized character’s body shape.
\textit{4) Naturalness}: Rank the images in order of the most natural-looking, considering factors such as physically impossible appearances, illogical features, inconsistencies, or lack of real-world physics and connections.
The study collected a total of 18,540 responses from 103 participants across 15 cases, including both custom and COCO-Wholebody scenarios. We present illustrative example images from the user study in Fig.~\ref{fig_supp_user_study_personalization}.

\noindent \textbf{Pose-Controlled Multi-Human Scene Generation}
We conducted a user study to assess the naturalness, face identity preservation, body shape preservation, and text-image correspondence of images generated by three baseline methods (one from each group) and our method. Participants ranked the top three methods based on the following criteria:

\textit{1) Text Correspondence}: Rank the images based on how closely they align with the textual description.
\textit{2) Pose Consistency}: Rank the images based on how well they reflect the given pose input.
\textit{3) Naturalness}: Rank the images in order of the most natural-looking, considering factors such as physically impossible appearances, illogical features, inconsistencies, or lack of real-world physics and connections.
The study collected a total of 15,390 responses from 114 participants across 15 cases, including both custom and COCO-Wholebody scenarios. We present illustrative example images from the user study in Fig.~\ref{fig_supp_user_study_pose}.
